\documentclass{article}
\usepackage{ijcai24}

\usepackage{times}
\usepackage{soul}
\usepackage{url}
\usepackage[hidelinks]{hyperref}
\usepackage[utf8]{inputenc}
\usepackage[small]{caption}
\usepackage{graphicx}
\usepackage{amsmath}
\usepackage{amsthm}
\usepackage{booktabs}
\usepackage{algorithm}
\usepackage{algorithmic}
\usepackage[switch]{lineno}

\usepackage{algorithm}
\usepackage{algorithmic}

\usepackage{newfloat}
\usepackage{listings}

\usepackage{amsmath}
\usepackage{amssymb}
\title{CLR-Face: Conditional Latent Refinement for Blind Face Restoration Using Score-Based Diffusion Models}

\author{
Maitreya Suin
\and
Rama Chellappa\\
\affiliations
Johns Hopkins University
}
\begin{document}

\maketitle

\begin{abstract}
    Recent generative-prior-based methods have shown promising blind face restoration performance. They usually project the degraded images to the latent space and then decode high-quality faces either by single-stage latent optimization or directly from the encoding. Generating fine-grained facial details faithful to inputs remains a challenging problem. Most existing methods produce either overly smooth outputs or alter the identity as they attempt to balance between generation and reconstruction. This may be attributed to the typical trade-off between quality and resolution in the latent space. If the latent space is highly compressed, the decoded output is more robust to degradations but shows worse fidelity. On the other hand, a more flexible latent space can capture intricate facial details better, but is extremely difficult to optimize for highly degraded faces using existing techniques. To address these issues, we introduce a diffusion-based-prior inside a VQGAN architecture that focuses on learning the distribution over uncorrupted latent embeddings. With such knowledge, we iteratively recover the clean embedding conditioning on the degraded counterpart. Furthermore, to ensure the reverse diffusion trajectory does not deviate from the underlying identity, we train a separate Identity Recovery Network and use its output to constrain the reverse diffusion process. Specifically, using a learnable latent mask, we add gradients from a face-recognition network to a subset of latent features that correlates with the finer identity-related details in the pixel space, leaving the other features untouched. Disentanglement between perception and fidelity in the latent space allows us to achieve the best of both worlds. We perform extensive evaluations on multiple real and synthetic datasets to validate the superiority of our approach.
\end{abstract}

\section{Introduction}
Blind face restoration (BFR) is a challenging problem that aims to recover a high-quality facial image from a low-quality input. With the increasing availability of degraded images captured in various real-world scenarios, the demand for effective BFR techniques has increased significantly in recent years. The primary goal of BFR is to recover undegraded facial features while preserving the person's identity. A complex combination of many factors, such as low resolution, blur, noise, compressions, etc., corrupts a facial image. Typically, traditional approaches \cite{baker2000hallucinating,5706362} rely on a degradation model and manually designed priors, which often lead to sub-optimal results and restricted ability to handle a variety of real-life images. Recently, the emphasis has shifted to deep learning-based approaches that leverage extensive datasets and achieve superior performance.
\newline Generative priors \cite{wang2021towards,yang2021gan} have shown remarkable improvement in restoration performance. Generative priors encapsulated in a well-trained high-quality face generator (e.g., StyleGAN \cite{karras2019style}) are typically exploited in such approaches. The degraded image is first projected to the latent space to get a `cleaner' latent vector either directly or after certain refinement operations. The decoder then projects the latent vector back to the pixel space, producing a restored image. Instead of dealing with a continuous latent space, \cite{gu2022vqfr,zhou2022towards} utilizes pretrained vector-quantized codebooks. The mechanism of vector quantization in a compressed latent space reduces the uncertainty of the task, making these methods robust to various degradations. Intuitively,  these approaches exploit information from two sources when restoring an image. The degraded input image contains crucial information about the person's identity, essential for faithful restoration. On the other hand, the pretrained decoder is utilized for a high-quality generation. Thus, ultimately, the task boils down to predicting the corresponding clean latent given a degraded image/latent. Using a highly compressed latent space usually reduces the complexity of the prediction task and prioritizes high-quality generation but fails to maintain intricate facial details essential for identity preservation. It may also lead to repeated and less diverse texture patterns in the output. On the other hand, using a less compressed latent space can be potentially more expressive and flexible but is often extremely difficult to optimize when directly projecting the degraded image to the latent space or trying to predict the correct latent code as a classification problem. A lower compression factor often diminishes the advantages of latent-space-based generative models, leading to suboptimal degradation removal from the input. Similar behavior was recently discussed in \cite{gu2022vqfr}, where the typical trade-off between fidelity and quality depending on the compression factor was analyzed. 
\begin{figure*}[h]
\centering
\includegraphics[width=0.85\textwidth]{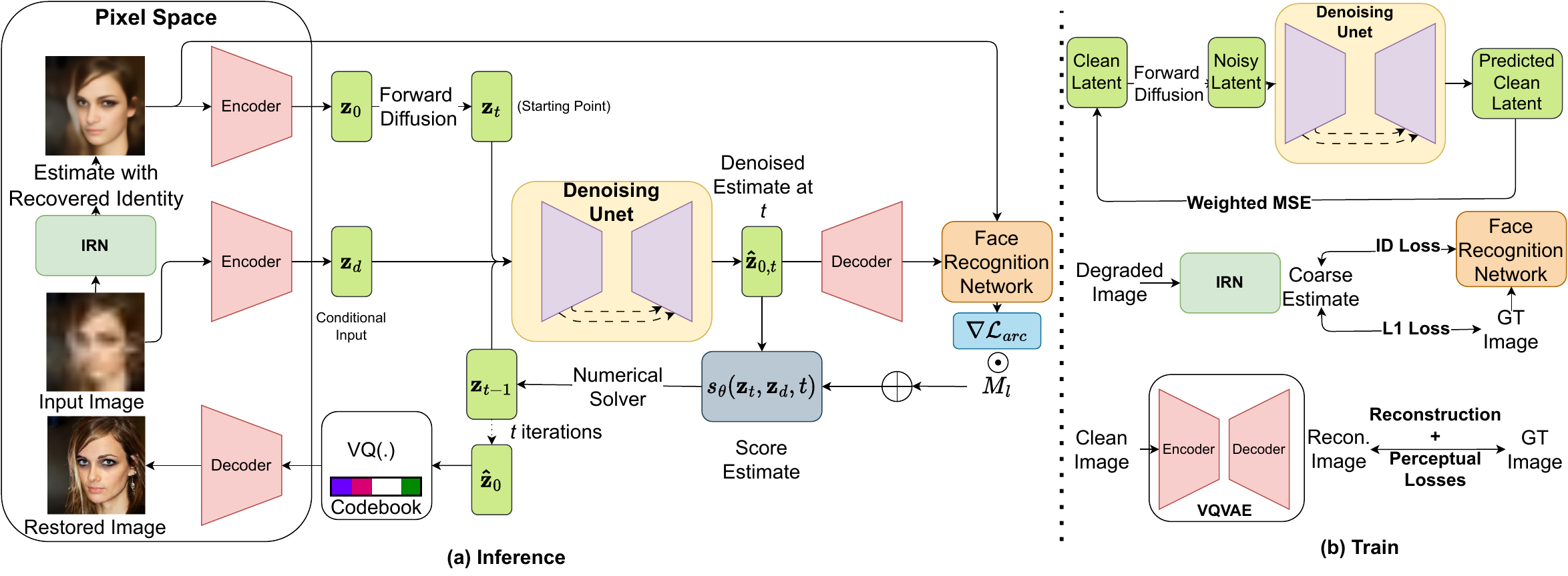}
\caption{An overview of our inference (left) and training framework (right).}
% \vspace{-5mm}
\label{fig:arch}
\end{figure*}
\newline We argue that the performance of such face restoration approaches mainly depends on the nature and size of the latent space. It is non-trivial to balance the demands of face generation and restoration using existing latent refinement techniques. \cite{gu2022vqfr,zhou2022towards} deploy a highly spatially-compressed latent space (downsampled by 32) and instead use skip-connections to improve the fidelity, which is often counter-productive for the output quality, as the information from the encoder is usually corrupted.  
\newline In this work, we design an alternative approach, where we first select a highly expressive latent space with lower compression and to handle its additional complexity, we introduce diffusion-based-prior to model it. We learn the clean latent space distribution by directly modeling the gradient of the log density of the data, known as the score function \cite{song2019generative}, using a neural network. Such gradient information can be utilized in reverse by stochastic sampling to generate diverse samples from the underlying distribution. Unlike unconditional generation as in \cite{song2020score}, for the current task, we model it as a conditional generation task, where given degraded latent, we perform multi-step refinement to produce the cleaner counterpart. 
\newline Our approach has several key advantages. 1. Similar to prior state-of-the-art (SOTA) blind face restoration works \cite{zhou2022towards}, we also utilize a pre-trained VQVAE framework to produce high-quality face images given correct latent codes. 2. The significantly lower compression ration allows us to maintain higher fidelity, suitable for BFR task. 3. Compared to the single-step refinement strategy of prior arts, multi-step conditional refinement of the corrupted latent using score-based diffusion prior is far more effective in producing the clean embedding, enabling us to use relatively mild compression rates, achieving more faithful and sharper reconstructions. 4. Compared to pixel-space-based diffusion models used in \cite{saharia2022image} (DDPM), our score-based latent-diffusion model has much better time complexity and requires significantly fewer refinement steps ($\approx 20 \times$). 
\newline The diffusion prior is extremely powerful for producing a clean latent. However, during multiple refinements, deviating from the trajectory of correct identity is plausible. To further constraint the refinement process, we deploy a separate Identity Recovery Network (IRN), which prioritizes recovering identity-specific intricacies given the input image using a face recognition loss. We utilize the output of the IRN to initialize the iterative refinement process. Next, at each step, we utilize a gradient-based score-update strategy using a pre-trained face recognition network to steer the refinement process along the identity recovered by the IRN. While this approach considerably diminishes the risk of identity distortion, the guidance signal from a face-recognition network may not align well with the other crucial objective of producing visually pleasing results. Such a dilemma is observed in existing works as well \cite{wang2021towards}, where usually a small amount of identity loss is introduced while prioritizing standard perceptual losses, which ultimately fails to recover the identity information adequately. Instead of blindly enforcing the identity constraint across the entire latent space, we introduce a learnable latent mask, where the gradient from the recognition network is harnessed to selectively update specific latent locations that contribute most significantly to identity information. This enables us to balance the restored identity and the perceptual quality derived from the diffusion prior. 
\newline To summarize, our main contributions are
\newline (1) We propose a diffusion-prior-based conditional latent refinement strategy for recovering the clean latent inside a pre-trained VQVAE framework.  
\newline (2) The strong modeling capability of such iterative refinement allows us to design a latent space with a lower spatial-compression ratio (f=4), improving the overall fidelity and sharpness of the output.
\newline (3) We further deploy an IRN to recover identity-specific features from the degraded input. The output of IRN, along with adaptive guidance using a learnable latent mask, steers the refinement process for better identity recovery without compromising the perceptual quality.
\newline (4) Extensive evaluation on multiple real and synthetic datasets demonstrates our proposed framework's superior perceptual and identity-preserving properties.

\section{Related Works}
Blind face restoration aims to recover a clear facial image from a degraded version without knowing the exact degradation. Various priors have been proposed, including geometric priors such as facial landmarks \cite{kim2019progressive,chen2018fsrnet,zhu2016deep} and parsing maps \cite{chen2021progressive,shen2018deep}, as well as reference-based approaches \cite{li2020blind,dogan2019exemplar}. However, these priors often struggle to estimate necessary information from corrupted images. Recent generative priors optimize the latent vector for GAN inversion techniques \cite{menon2020pulse,gu2020image} or direct projection of the input image to the latent space \cite{richardson2021encoding}. \cite{yang2021gan}, and \cite{wang2021towards} exploited the generative prior inside an encoder-decoder framework, with structural details from the degraded input through skip connections. For severe degradations, such links lead to unwanted artifacts. 
\newline Diffusion and score-based models have shown improvements recently \cite{saharia2022image,whang2022deblurring,choi2021ilvr,yue2022difface,chung2022come}. An iterative refinement strategy that repeats the reverse process several times has been adopted by \cite{saharia2022image}, \cite{whang2022deblurring} for super-resolution and motion deblurring tasks. \cite{choi2021ilvr} used a pre-trained diffusion model and guided the reverse process with low-frequency information from a conditional image. However, such a conditioning strategy does not translate well for the BFR problem with significant degradation and may alter the identity for many steps. \cite{yue2022difface} uses an unconditional diffusion model and starts from an intermediate stage of the reverse diffusion process. But, as the underlying diffusion model is unconditional, the restored face changes considerably compared to the original person if the reverse process is run longer. If it is used for a smaller timespan, the visual quality and sharpness of the output suffer considerably. \cite{chung2022come} addressed only non-blind super-resolution tasks, and its identity-preserving capability is yet to be tested for more difficult blind face restoration scenarios. Our work mainly aims to balance identity preservation and perceptual quality for heavily degraded faces.

\section{Method}
Our framework has two main parts, a Vector-Quantized (VQ) autoencoder is our backbone that maps a degraded image to the latent space with a downsampling factor $f=4$ and reverses this operation after refining the latent. A conditional diffusion model (denoising UNet) is used for the latent refinement task. We pre-train the quantized autoencoder through self-reconstruction to obtain a discrete codebook and the corresponding decoder for projecting the restored latent back to pixel space. In addition, we also deploy IRN to produce an initial estimate of the restored face, focusing on recovering identity-specific features, which are used to steer the iterative latent refinement process. An overview of our framework is shown in Fig. \ref{fig:arch}. In the following sections, we describe the details of different parts of our framework.

\subsection{Learning Latent Space of Images via Vector-Quantized Codebook}
We use $x$ and $\mathbf{z}$ to represent image in pixel space and the corresponding latent representation (extracted using the encoder of VQVAE), respectively. Our perceptual compression model utilizes a variant of the Vector-Quantize codebook called VQGAN \cite{esser2021taming}. It learns a perceptual codebook through a combination of perceptual loss \cite{johnson2016perceptual} and adversarial training objectives \cite{isola2017image}, preventing blurriness that can occur when relying solely on pixel-space $L_2$ or $L_1$ losses. The model consists of an encoder $E$, a decoder $G$, and a codebook $\mathcal{Z} = {z_k}_{k=1}^{K} \in \mathbb{R}^{K \times d}$ containing a finite set of embedding vectors. Here, $K$ represents the codebook size, and $d$ denotes the dimension of the codes. Given an input image $x \in \mathbb{R}^{3 \times H \times W}$, we project it to the latent space as $\textbf{z} = E(x) \in \mathbb{R}^{d \times \frac{H}{f} \times \frac{W}{f}}$. Next, a spatial collection of image tokens $\textbf{z}_q$ is obtained using a spatial-wise quantizer $Q(\cdot)$, which maps each spatial feature $z_{ij}$ into its closest codebook entry $z_k$
\begin{equation}
    \textbf{z}_q = Q(\textbf{z}) = \left(\operatorname*{argmin}_{z_k \in \mathcal{Z}} ||z_{ij} - z_k||_2^2 \right) \in \mathbb{R}^{d \times \frac{H}{f} \times \frac{W}{f}}
\end{equation}
The decoder reconstructs the image from the latent, giving $\tilde{x} = G(z_q)$. The encoder $E$, the decoder $G$ and the codebook $\mathcal{Z}$ are trained end-to-end via the following loss function
\begin{multline}
    \mathcal{L} = ||x - \tilde{x}||_1 + ||\phi(x) - \phi(\tilde{x})||_1 + [\log D(x) + \log (1-D(\tilde{x}))] + \\ ||\text{sg}(E(x)) - \textbf{z}_q||_2^2 + ||\text{sg}(\textbf{z}_q) - E(x)||_2^2
\end{multline}
where $\text{sg}(\cdot)$ stands for the stop-gradient operator, $\phi(\cdot)$ represents the VGG-based \cite{simonyan2014very} feature extractor. The first three terms help reconstruct the input image, whereas the last two terms reduce the distance between codebook $\mathcal{Z}$ and input feature embeddings $\textbf{z}$. As we aim to build upon the already established foundation of continuous space diffusion models, we absorb the quantization layer in the decoder. Given $\textbf{z}$, we use the conditional diffusion model to recover the correct latent embedding through iterative refinement. Next, quantization and the pre-trained high-quality codebook can handle any inaccuracies in the latent prediction for producing a clean, restored face.

Compared to existing BFR works that use VQVAE \cite{gu2022vqfr,zhou2022towards}, our design has two critical differences: we use a milder compression (downsample $f=4$ instead $32$) and a larger codebook size ($K=8192$ instead of $1024$). As demonstrated in the experimental section, such an expressive yet highly complex latent space can be optimally modeled by iterative diffusion-prior-based refinement. As a result, the restored image is of higher fidelity and realness than prior art.

\subsection{Iterative Latent Refinement Using Diffusion Prior}
Given the high dimensional latent with a lower compression ratio, existing methods encounter challenges in generating a cleaner latent through a single pass in the encoder or a classification-based refinement module, as elaborated in \cite{gu2022vqfr}. Instead, we aim to learn the probability distribution of the clean latent space (through gradients of log probability density functions) and map the degraded latent to the clean space in multiple refinement steps. Our approach is built upon the framework of \cite{song2019generative}, where stochastic differential equations drive the underlying diffusion process. We can represent this process using continuous time variables ${\textbf{z}_t}_{t=0}^{t=1}$, where $\textbf{z}_0$ represents the initial variable (i.e., clean latent representation) and $\textbf{z}_t$ represents its noisy version at time $t$.  The diffusion process is defined by an Itô SDE \cite{rogers2000diffusions}
\begin{equation}
    \label{eq:fwd_diff}
    d\textbf{z}_t = f(\textbf{z}_t, t), dt + g(t) d\textbf{w}_t
\end{equation}
where $f(\cdot,\cdot)$ and $g(\cdot)$ are the drift and diffusion coefficients, respectively, and $\textbf{w}_t$ denotes the standard Brownian motion. We denote the distribution of $\textbf{z}_t$ as $p(\textbf{z}_t)$. 

The SDE in Eq. \ref{eq:fwd_diff} can be converted to a generative model by first sampling from $\textbf{z}_1 \sim  \mathcal{N}(z_1, 0, I)$ and then running the
reverse-time SDE as
\begin{equation}
\label{eq:rev_diff_sde}
    d\textbf{z}_t = [f(\textbf{z}_t, t) - \frac{1}{2}g^2(t) \nabla_{\textbf{z}} \log p(\textbf{z}_t)]dt +  g(t) d\bar{\textbf{{w}}}_t
\end{equation}
where $\bar{\textbf{{w}}}_t$ is a reverse-time standard Wiener process and $dt$ is an infinitesimal negative time step. 
 
As shown in \cite{song2020score}, there exists a deterministic ordinary differential equation (ODE), whose trajectories share the same marginal probability densities as the SDE (Eqs. \ref{eq:fwd_diff}, \ref{eq:rev_diff_sde}). It can be described as
\begin{equation}
    \label{eq:rev_diff_ode}
    d\textbf{z}_t = [f(\textbf{z}_t, t) - g^2(t) \nabla_{\textbf{z}} \log p(\textbf{z}_t)]dt
\end{equation}
In practice, blackbox ODE solvers can be used on Eq. \ref{eq:rev_diff_ode} for faster and high quality sampling, assuming we know the score function. The score function ($\nabla_{\textbf{z}} \log p(\textbf{z}_t)$) is typically approximated using a neural network $s_\theta(\textbf{z}_t, t)$ \cite{vincent2011connection,song2019generative,ho2020denoising}, trained to minimize the weighted Fisher’s divergence with a positive weighting function $\lambda$ as
\begin{equation}
    \label{eq:score_matching_1}
    \min_\theta \mathbb{E}_{t \sim \mathbb{U}[0,1]}(\lambda(t) ||\nabla_{\textbf{z}} \log p(\textbf{z}_t) -  s_\theta(\textbf{z}_t, t)||_2^2)
\end{equation}
\subsubsection{Conditional Diffusion for BFR}
Instead of unconditionally sampling from the underlying clean latent distribution using Eq. \ref{eq:rev_diff_ode}, for the task of BFR, our goal is to predict the clean counterpart, given a degraded latent $\textbf{z}^{d}$ of the input image $\textbf{x}^{d}$. We are interested in $p(\textbf{z}_0|\textbf{z}^d)$, where $\textbf{z}_0$ is the target clean latent distribution corresponding to the image $\textbf{x}^d$ and $\textbf{z}^d$ is the condition signal. Similar to Eq. \ref{eq:rev_diff_ode}, we can obtain a conditional reverse-time ODE as
\begin{equation}
    \label{eq:rev_diff_ode_cond}
    d\textbf{z}_t = [f(\textbf{z}_t, t) - g^2(t) \nabla_{\textbf{z}} \log p(\textbf{z}_t|\textbf{z}^d)]dt
\end{equation}
We need to learn the conditional score function $\nabla_\textbf{z} \log p(\textbf{z}_t|\textbf{z}^d)$ in order to be able to sample from Eq. \ref{eq:rev_diff_ode_cond}  using reverse-time diffusion. In \cite{song2020score}, $\nabla_\textbf{z} \log p(\textbf{z}_t|\textbf{z}^d)$ was estimated using
\begin{equation}
    \nabla_\textbf{z} \log p(\textbf{z}_t|\textbf{z}^d) =  \nabla_{\textbf{z}} \log p(\textbf{z}_t) + \nabla_{\textbf{z}} \log p(\textbf{z}^d|\textbf{z}_t)
\end{equation}
where $\nabla_\textbf{z} \log p(\textbf{z}_t)$ is learned using an unconditional model and $\nabla_\textbf{z} \log p(\textbf{z}^d|\textbf{z}_t)$ is learned using a separate auxiliary network. But, modeling such a forward process is non-trivial for BFR, as there could be many possible $\textbf{z}^d$ given a $\textbf{z}_t$. Instead, following the intuition of \cite{saharia2022image,tashiro2021csdi}, we can extend the formulation of Eq. \ref{eq:score_matching_1} to a conditional case as
\begin{equation}
    \label{eq:score_matching_1_cond}
    \min_\theta \mathbb{E}_{{\substack{t \sim \mathbb{U}[0,1] \\  \textbf{z}^d \sim p(\textbf{z}^d)}}
}(\lambda(t) ||\nabla_{z} \log p(\textbf{z}_t)] -  s_\theta(\textbf{z}_t, \textbf{z}^d, t)||_2^2)
\end{equation}
where we pass the condition $\textbf{z}^d$ directly to the denoising UNet, making it learn the conditional score. We empirically verify that it works well for BFR task. In \cite{song2020score,kadkhodaie2020solving}, the noise was added to both the target and the conditional observations, which we found to be harmful. We follow the simplified configuration of \cite{karras2022elucidating}, where we set $f(\textbf{z}_t, t) = 0$ and $g(t) = \sqrt{2}t$. Then, $p(\textbf{z}_t)$ can be expressed as $p(\textbf{z}_0) * \mathcal{N}(0, t^2 I)$, where $*$ represents convolution operation and $p(\textbf{z}_0)$ is the underlying distribution of the clean or non-degraded images' latent space, which we aim to model. Once we know the score function (using a neural network), Eq. \ref{eq:rev_diff_ode_cond} can be re-written as
\begin{equation}
    \label{eq:rev_diff_ode_simple}
    d\textbf{z}_t = -t s_\theta(\textbf{z}_t, \textbf{z}^d, t)
\end{equation}
During the reverse process, we can proceed backward in time with a black box ODE solver, such as Heun solvers \cite{karras2022elucidating}, to obtain the solution trajectory. The resulting $\hat{\textbf{z}}_0$ can be considered an approximate embedding of the clean image corresponding to $x^{d}$. $\hat{\textbf{z}}_0$ is passed through the pre-trained decoder to produce the restored image. The decoder has the quantization layer, which handles minor errors in the continuous space prediction.
\begin{figure*}[t]
\setlength{\tabcolsep}{1pt}
\scriptsize
\centering
%\hspace{-0.4cm}
\begin{tabular}{ccccccccc}
\includegraphics[width=0.095\linewidth]{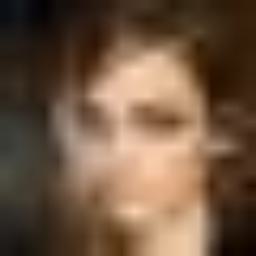} &
\includegraphics[width=0.095\linewidth]{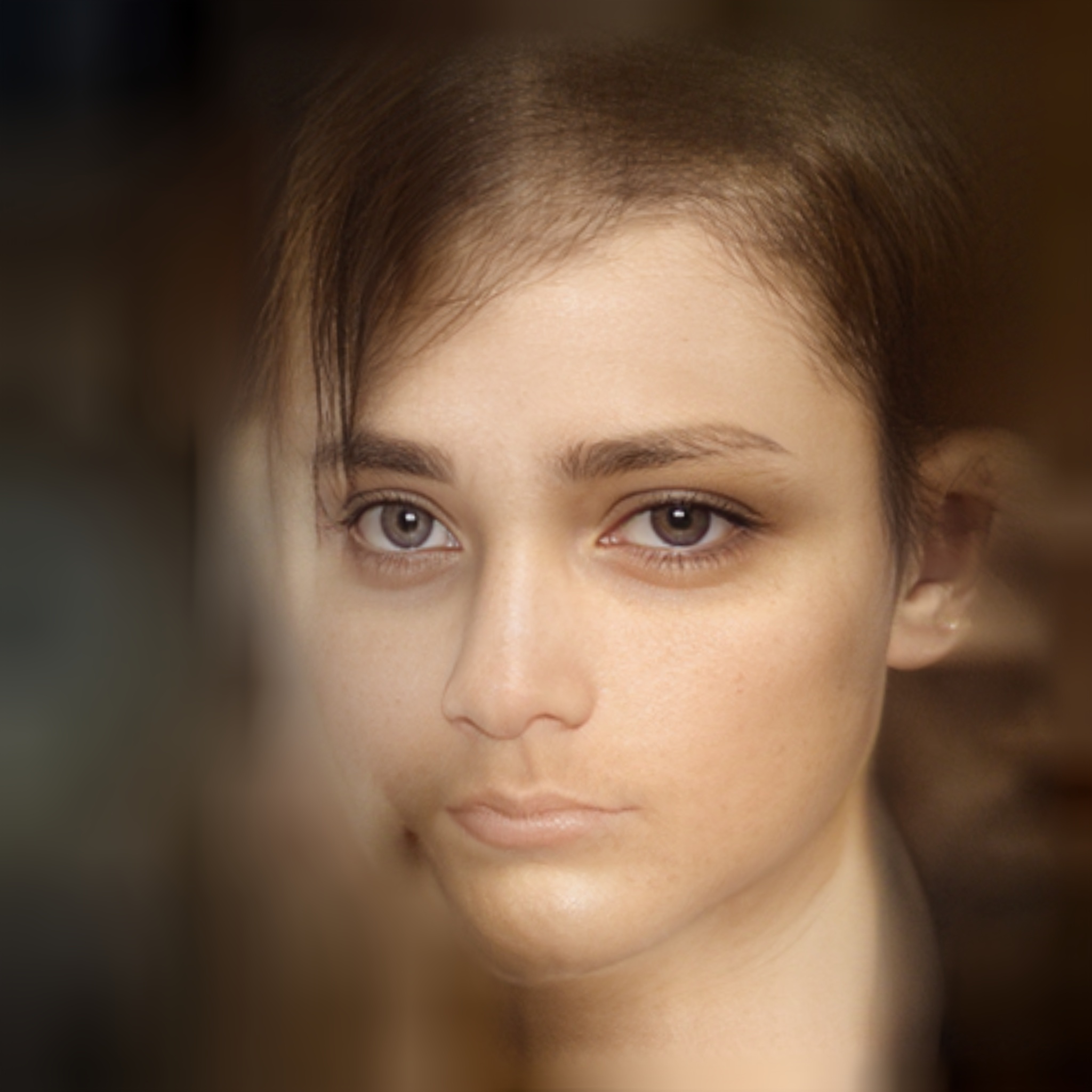} &
\includegraphics[width=0.095\linewidth]{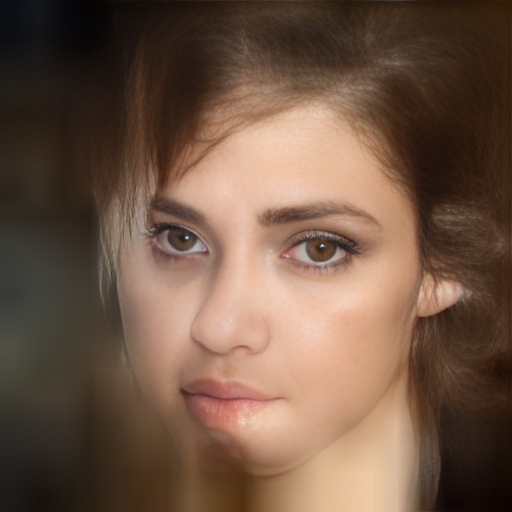} &
\includegraphics[width=0.095\linewidth]{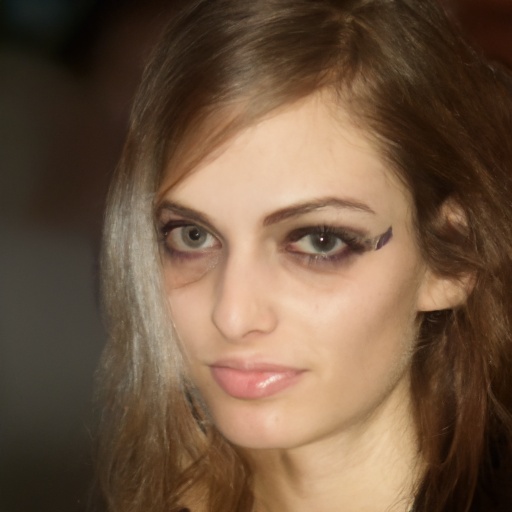} &
\includegraphics[width=0.095\linewidth]{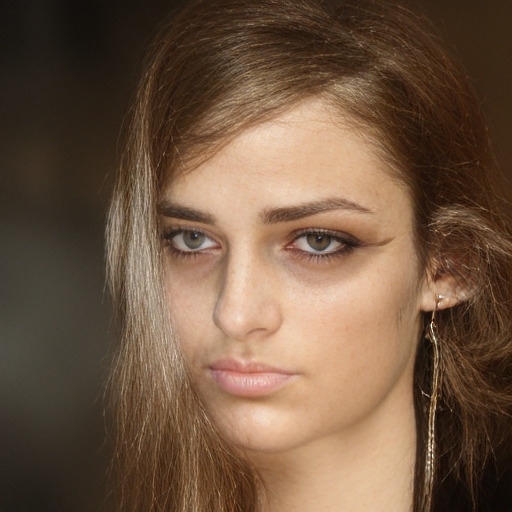} &
\includegraphics[width=0.095\linewidth]{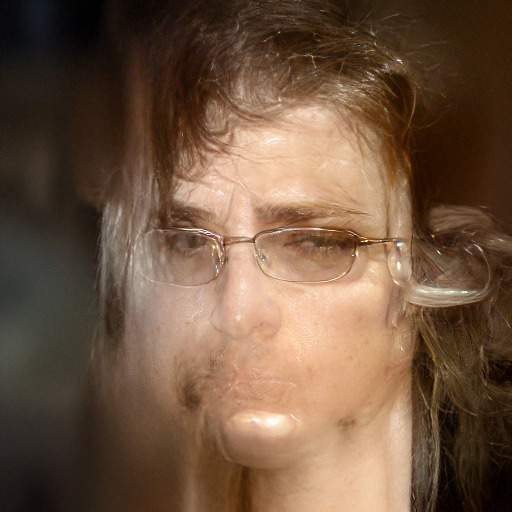} &
\includegraphics[width=0.095\linewidth]{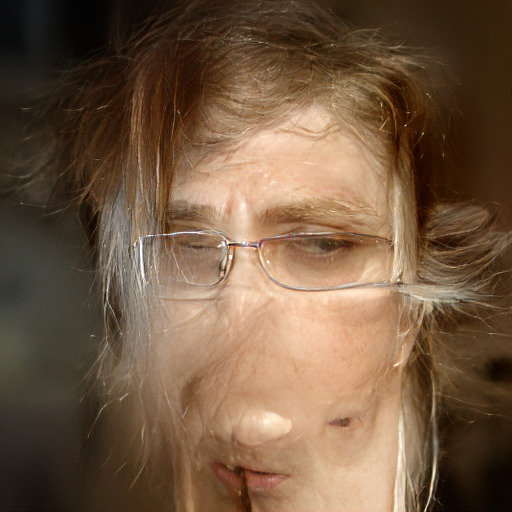} &
\includegraphics[width=0.095\linewidth]{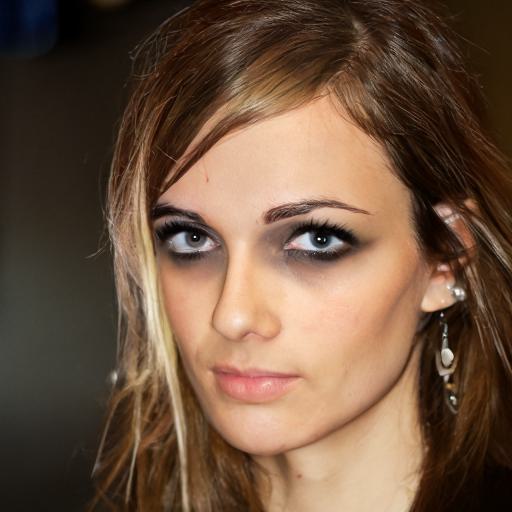} &
\includegraphics[width=0.095\linewidth]{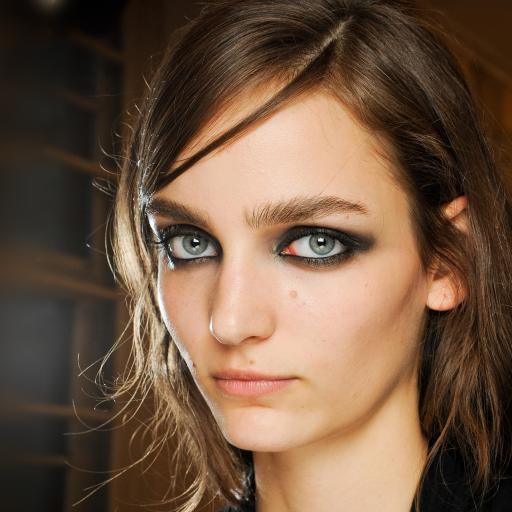}
\\
\includegraphics[width=0.095\linewidth]{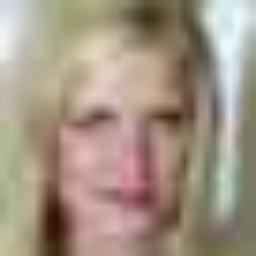} &
\includegraphics[width=0.095\linewidth]{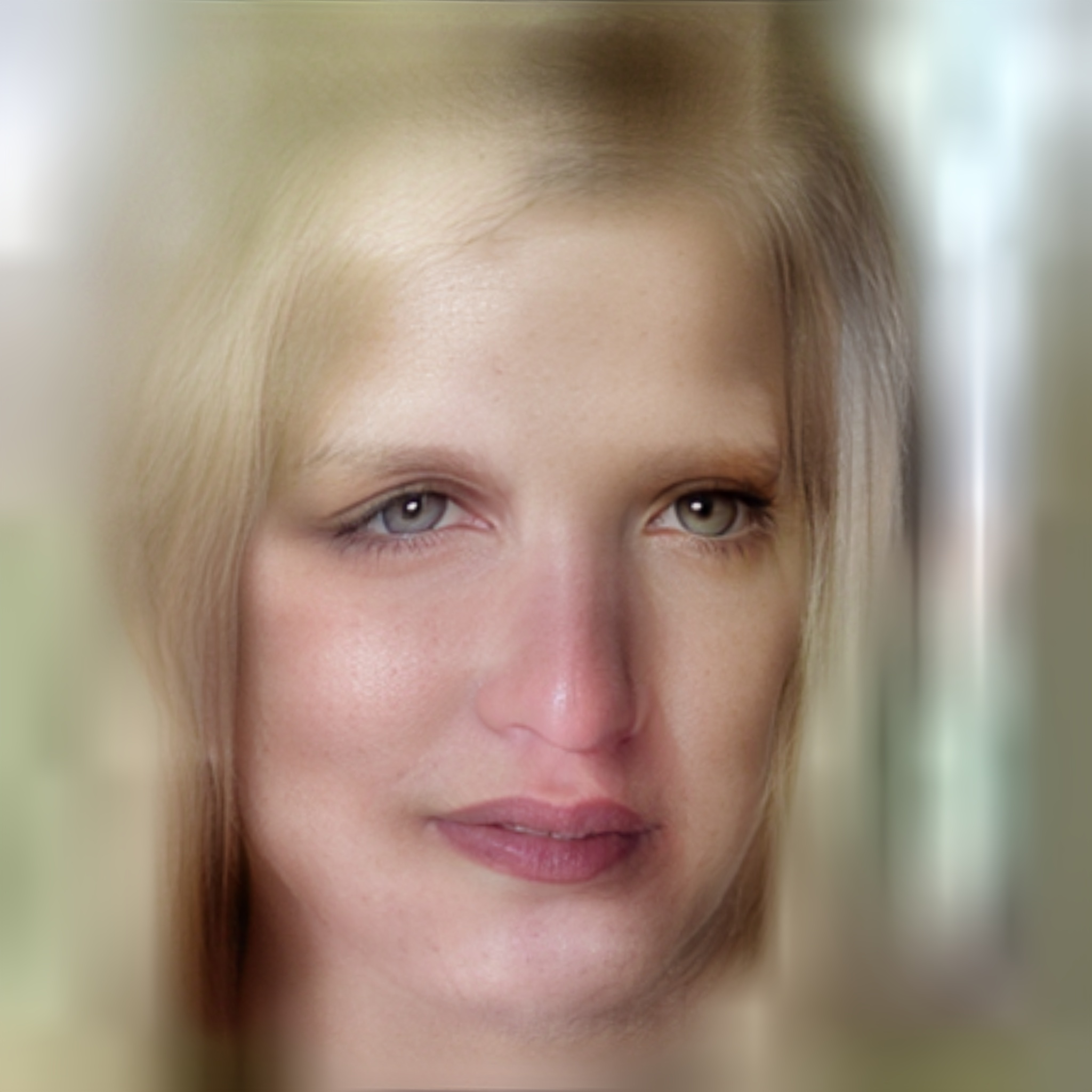} &
\includegraphics[width=0.095\linewidth]{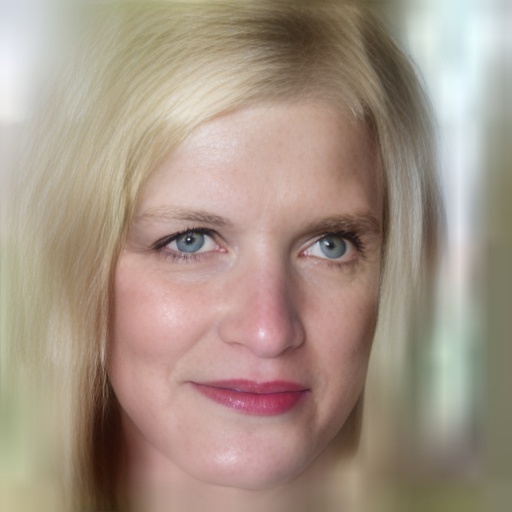} &
\includegraphics[width=0.095\linewidth]{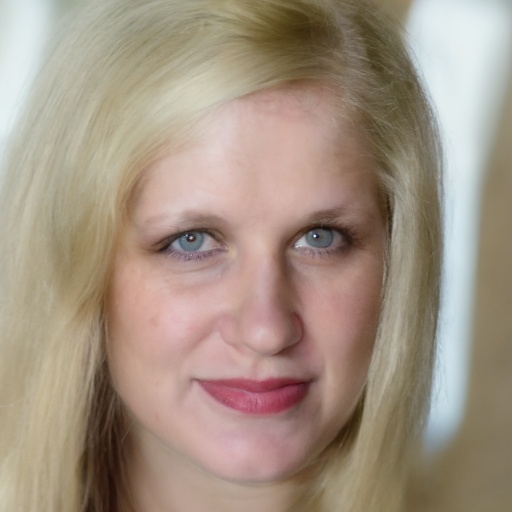} &
\includegraphics[width=0.095\linewidth]{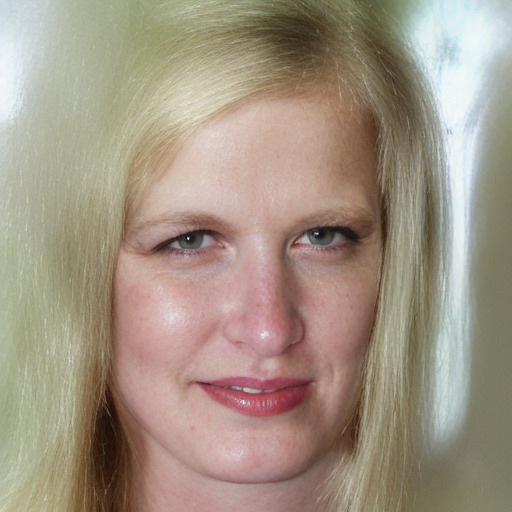} &
\includegraphics[width=0.095\linewidth]{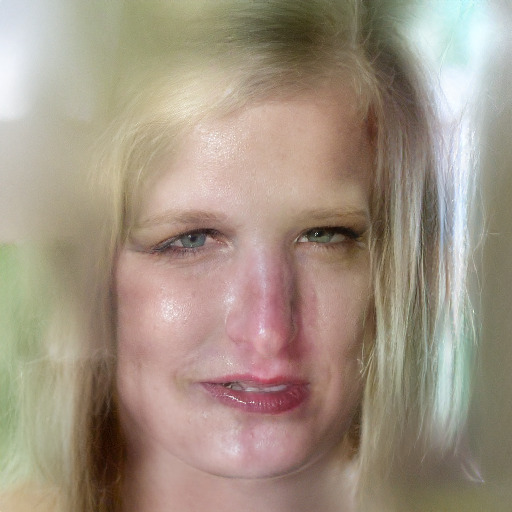} &
\includegraphics[width=0.095\linewidth]{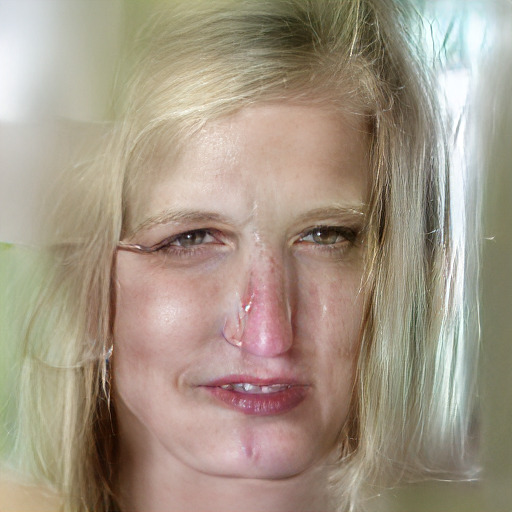} &
\includegraphics[width=0.095\linewidth]{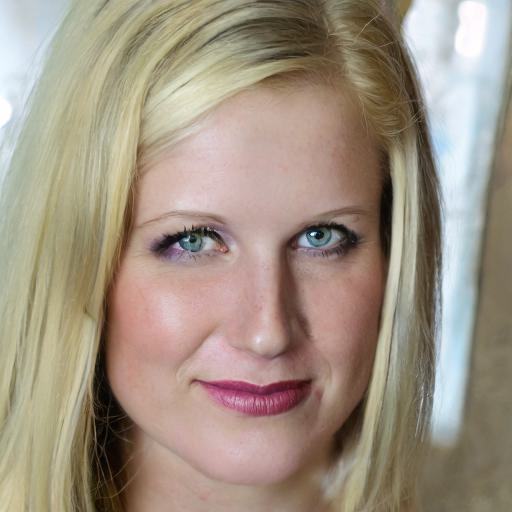} &
\includegraphics[width=0.095\linewidth]{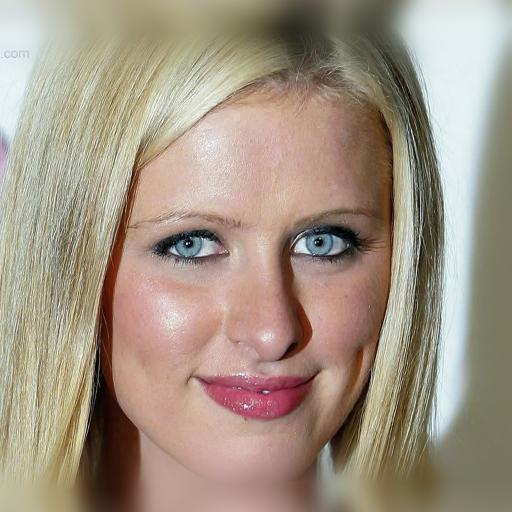}
\\
Input & GPEN & GFPGAN & DifFace & CodeFormer & RestoreFormer & VQFR & Ours & GT
\end{tabular}\\ %\hspace{-2.3mm}

\caption{Qualitative comparisons on CelebA-Test set for BFR.}
\vspace{-2mm}
\label{fig:celeba_bfr}
\end{figure*}

\begin{figure*}[t]
\setlength{\tabcolsep}{1pt}
\scriptsize
\centering
%\hspace{-0.4cm}
\begin{tabular}{ccccccccc}
\includegraphics[width=0.095\linewidth]{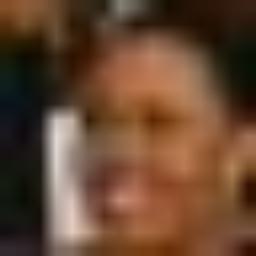} &
\includegraphics[width=0.095\linewidth]{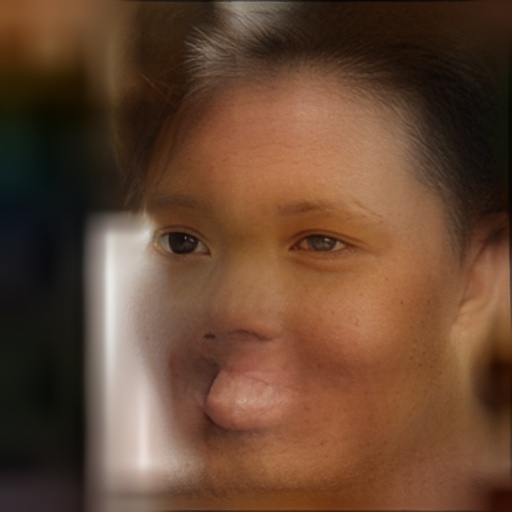} &
\includegraphics[width=0.095\linewidth]{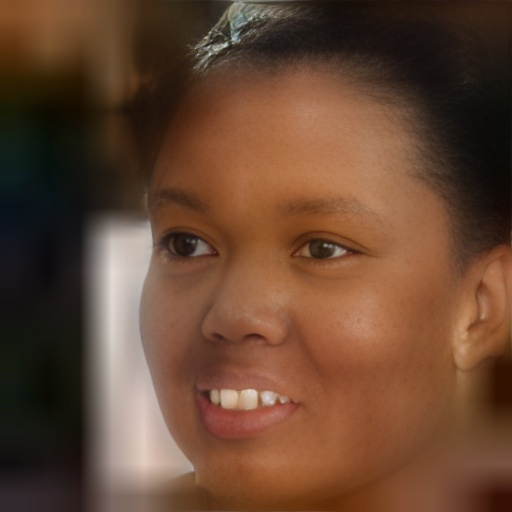} &
\includegraphics[width=0.095\linewidth]{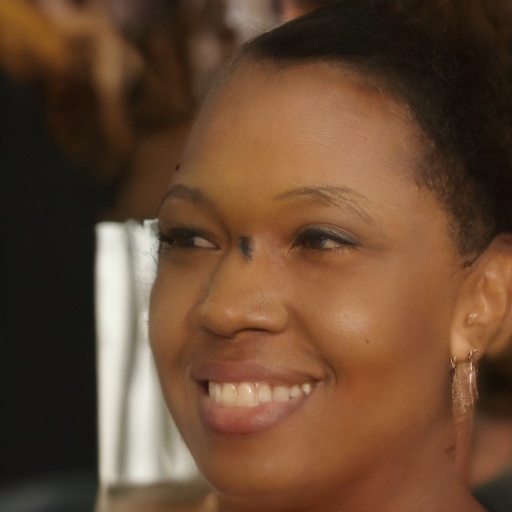} &
\includegraphics[width=0.095\linewidth]{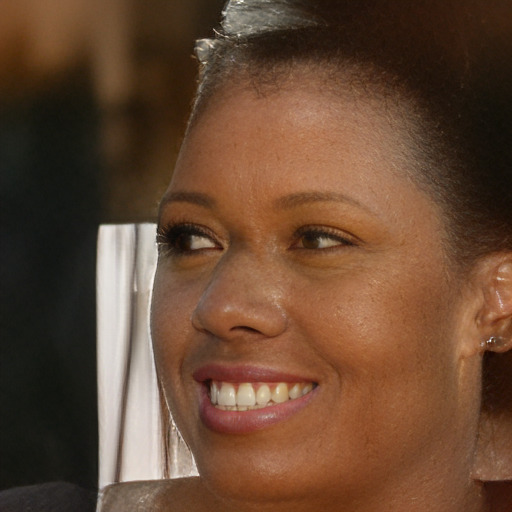} &
\includegraphics[width=0.095\linewidth]{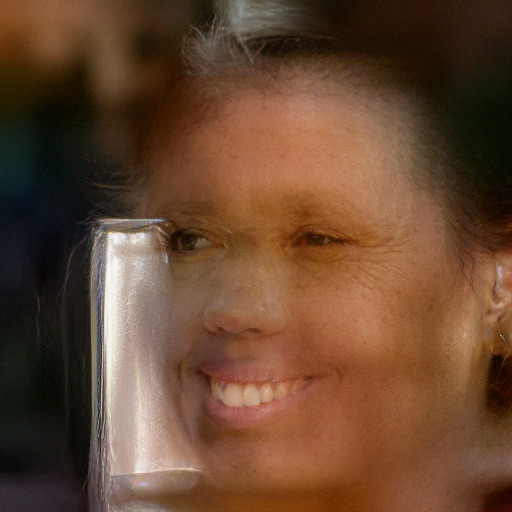} &
\includegraphics[width=0.095\linewidth]{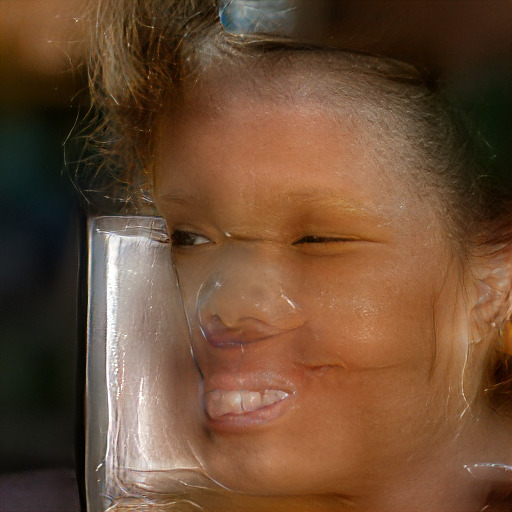} &
\includegraphics[width=0.095\linewidth]{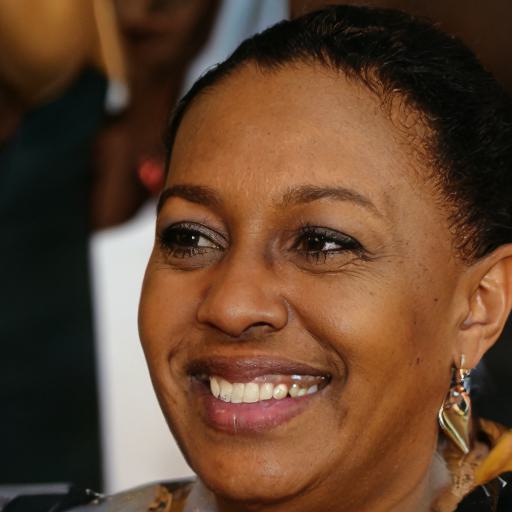} &
\includegraphics[width=0.095\linewidth]{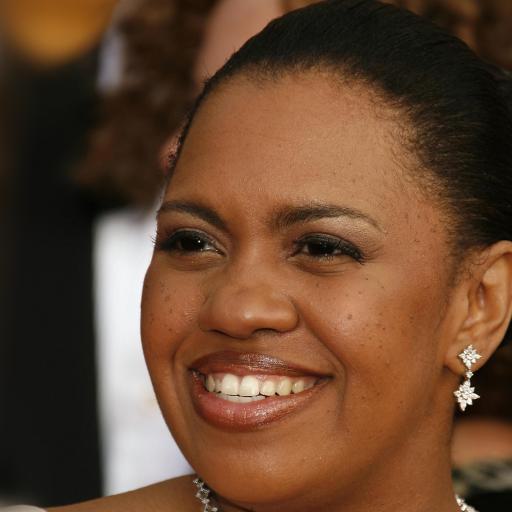}
\\
\includegraphics[width=0.095\linewidth]{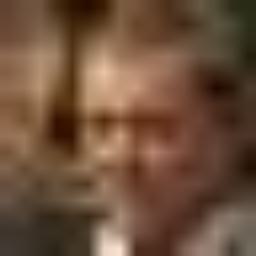} &
\includegraphics[width=0.095\linewidth]{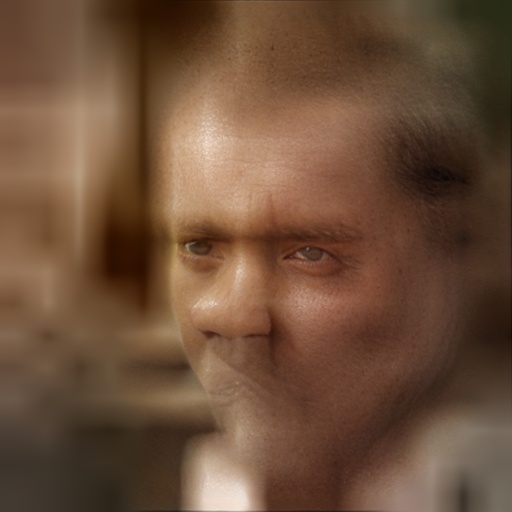} &
\includegraphics[width=0.095\linewidth]{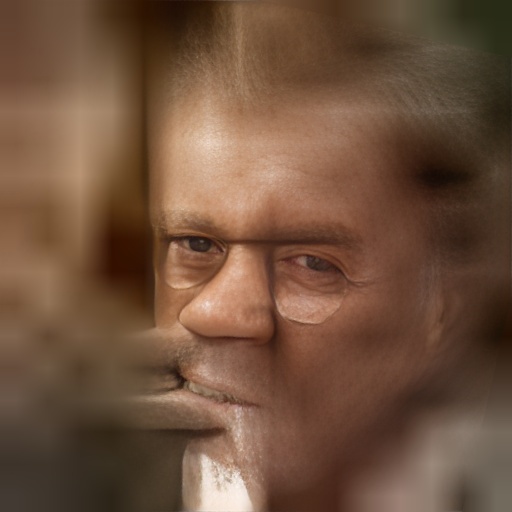} &
\includegraphics[width=0.095\linewidth]{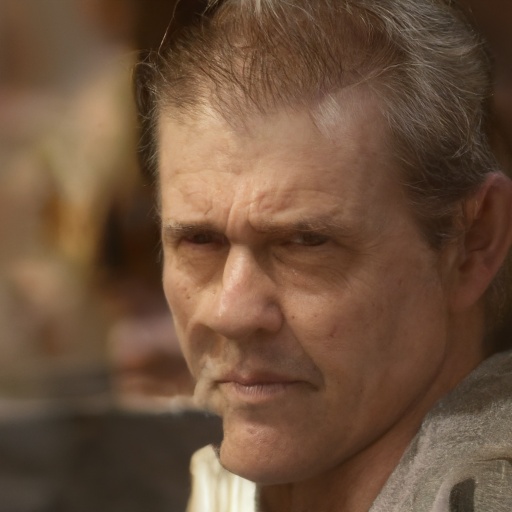} &
\includegraphics[width=0.095\linewidth]{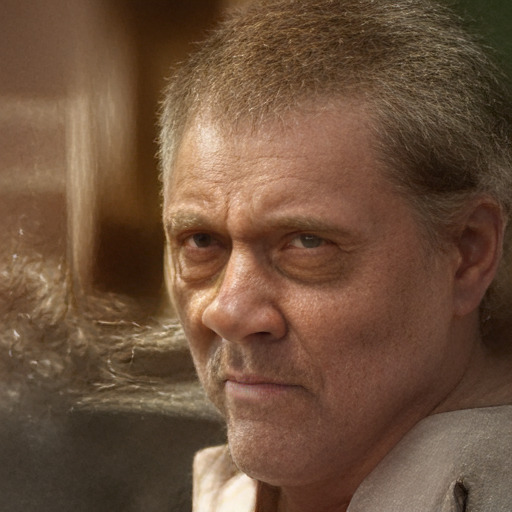} &
\includegraphics[width=0.095\linewidth]{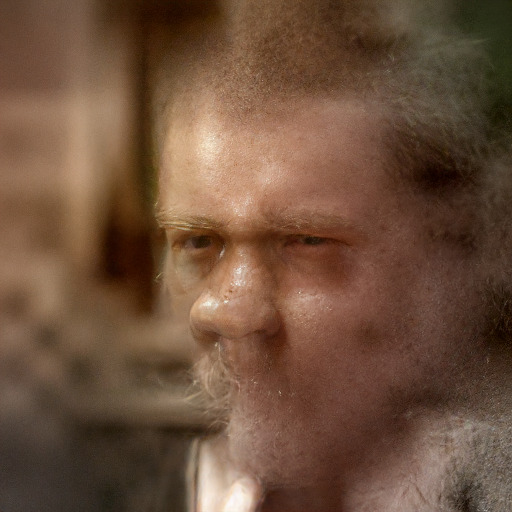} &
\includegraphics[width=0.095\linewidth]{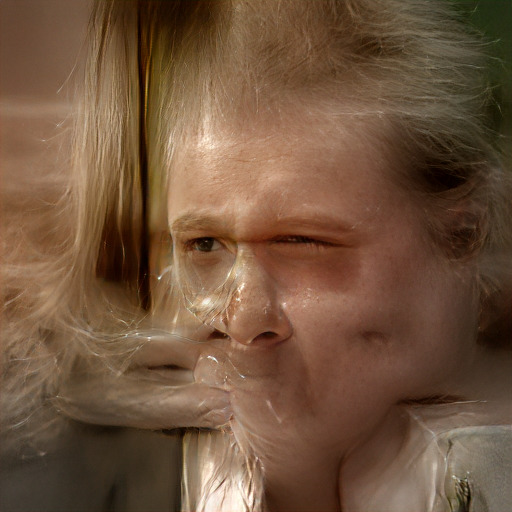} &
\includegraphics[width=0.095\linewidth]{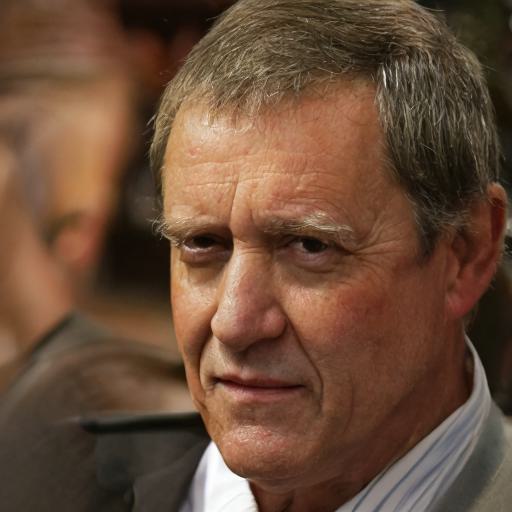} &
\includegraphics[width=0.095\linewidth]{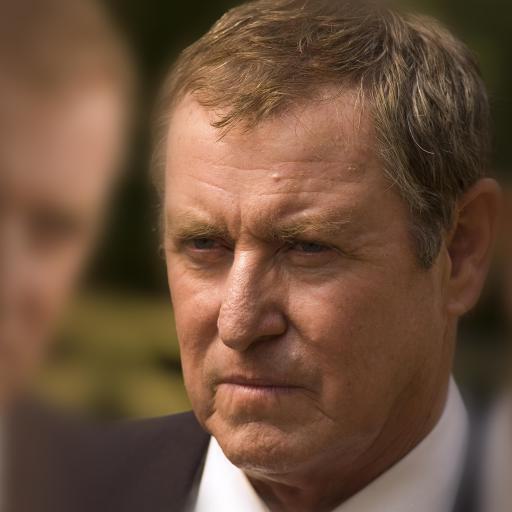}

\\
Input & GPEN & GFPGAN & DifFace & CodeFormer & RestoreFormer & VQFR & Ours & GT
\end{tabular}\\ %\hspace{-2.3mm}

\caption{Qualitative comparisons on CelebA-Test set for $\times 32$ upsampling. Although the input is severely degraded, our approach works better than existing works in restoring the face faithfully.}
\label{fig:celeba_sr}
\vspace{-3mm}
\end{figure*}

\begin{figure*}[t]
\setlength{\tabcolsep}{1pt}
\scriptsize
\centering
%\hspace{-0.4cm}
\begin{tabular}{ccccccccc}
\includegraphics[width=0.1\linewidth]{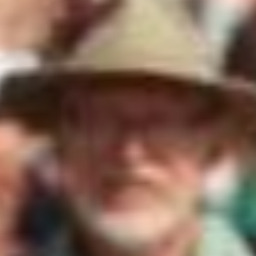} &
\includegraphics[width=0.1\linewidth]{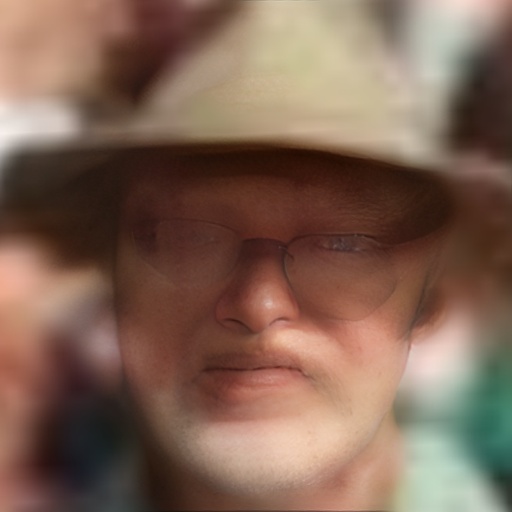} &
\includegraphics[width=0.1\linewidth]{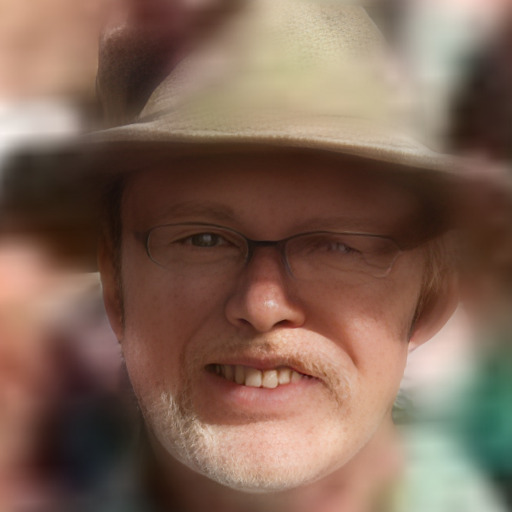} &
\includegraphics[width=0.1\linewidth]{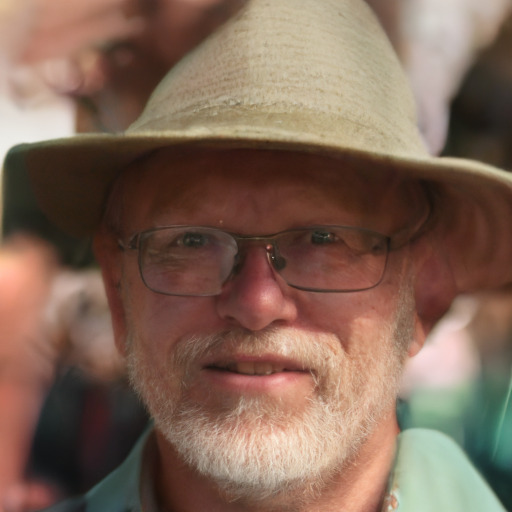} &
\includegraphics[width=0.1\linewidth]{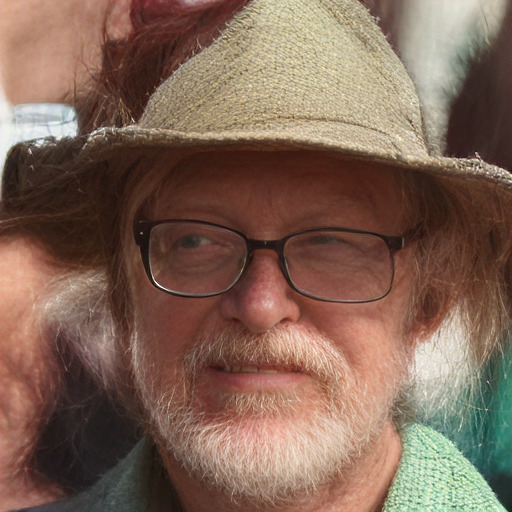} &
\includegraphics[width=0.1\linewidth]{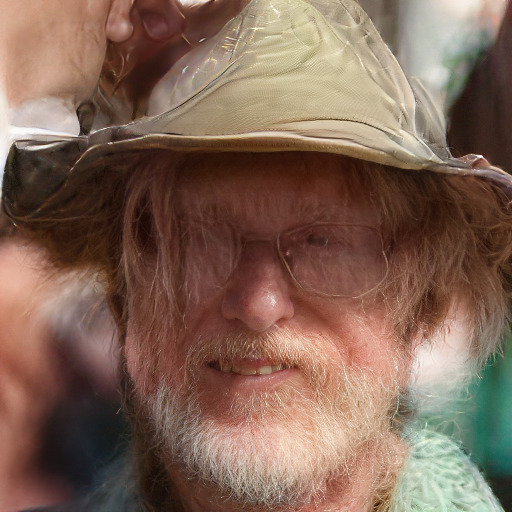} &
\includegraphics[width=0.1\linewidth]{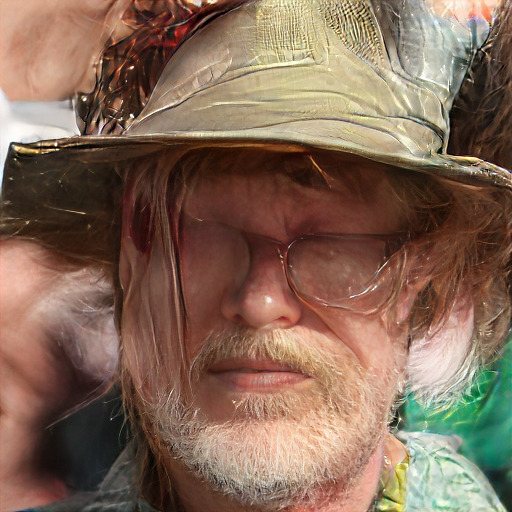} &
\includegraphics[width=0.1\linewidth]{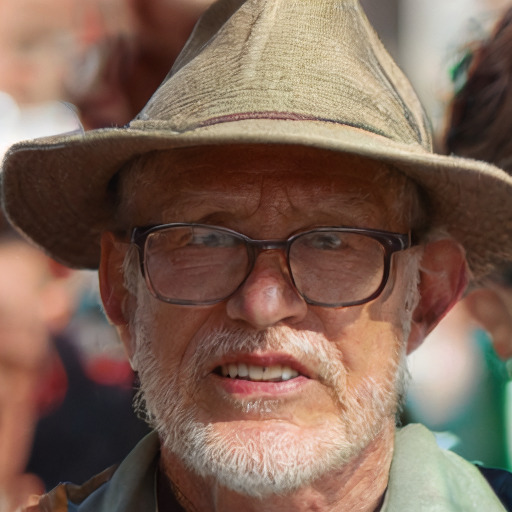} 
\\
\includegraphics[width=0.1\linewidth]{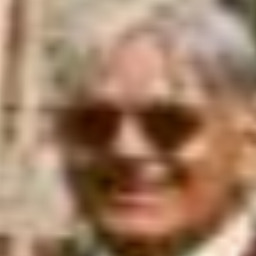} &
\includegraphics[width=0.1\linewidth]{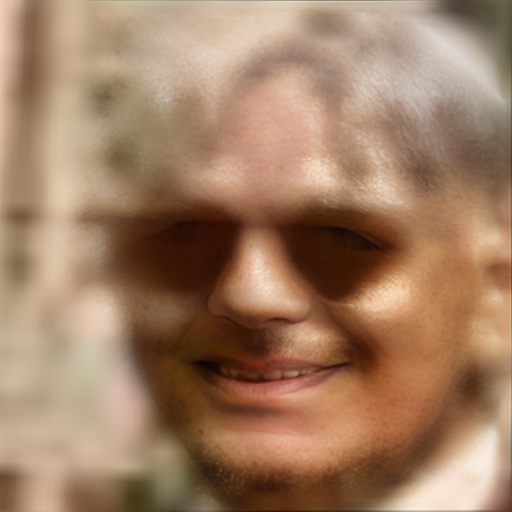} &
\includegraphics[width=0.1\linewidth]{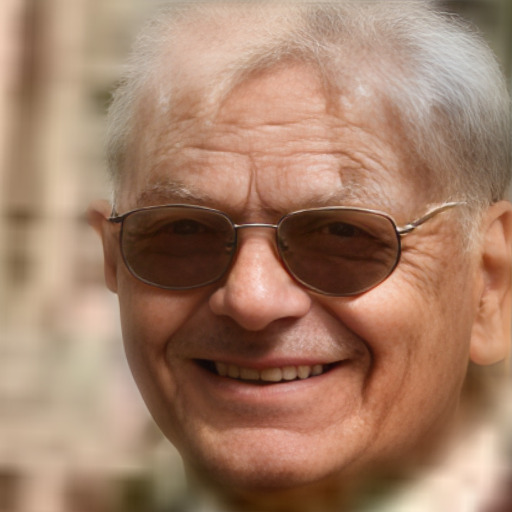} &
\includegraphics[width=0.1\linewidth]{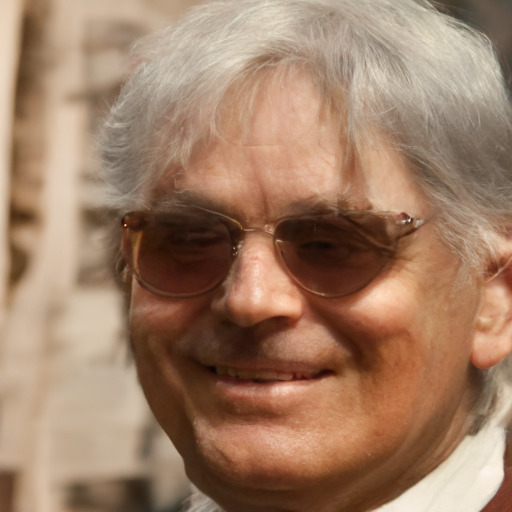} &
\includegraphics[width=0.1\linewidth]{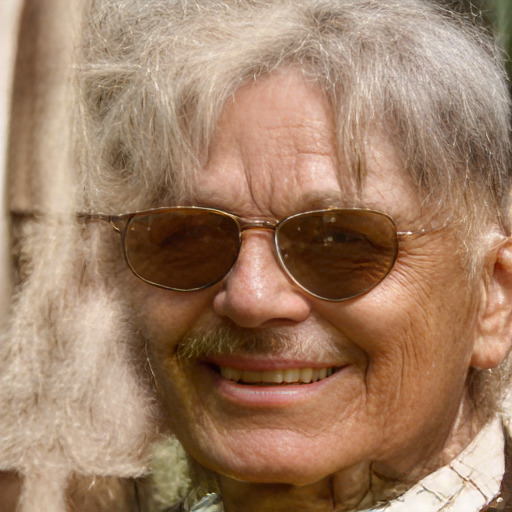} &
\includegraphics[width=0.1\linewidth]{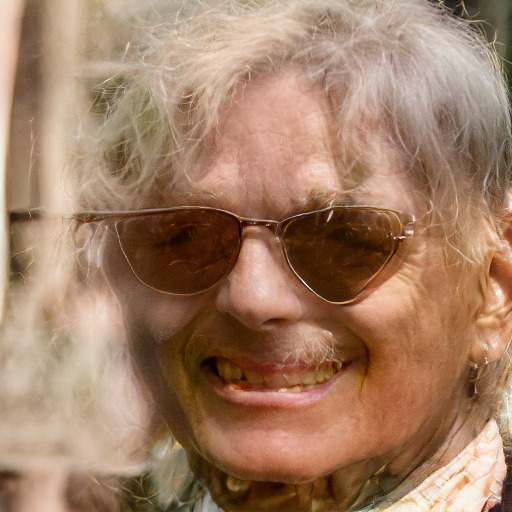} &
\includegraphics[width=0.1\linewidth]{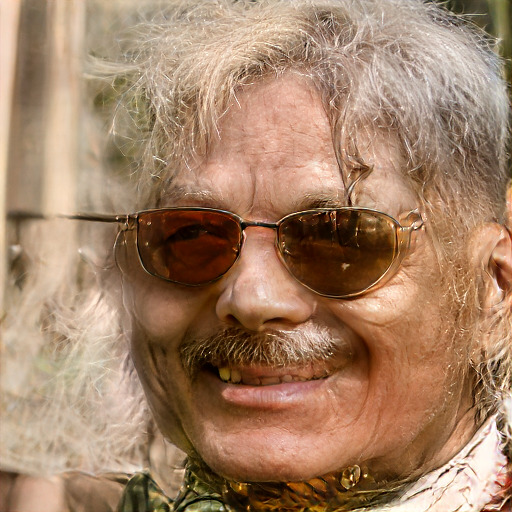} &
\includegraphics[width=0.1\linewidth]{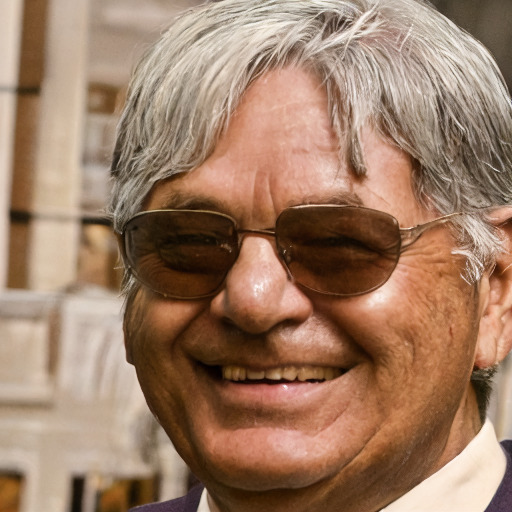} 
\\
\includegraphics[width=0.1\linewidth]{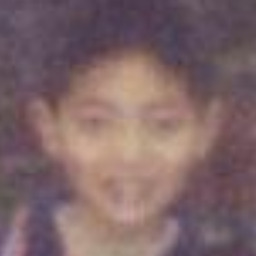} &
\includegraphics[width=0.1\linewidth]{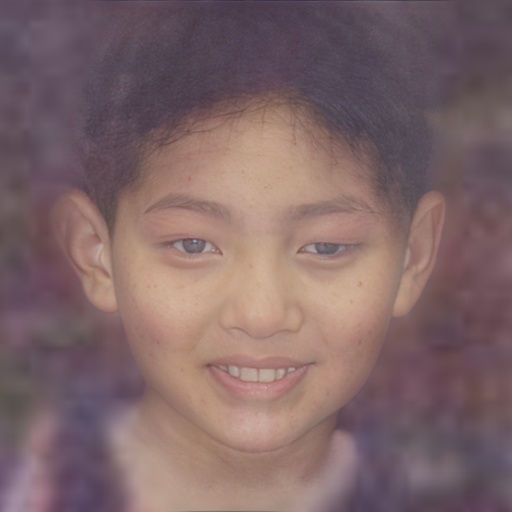} &
\includegraphics[width=0.1\linewidth]{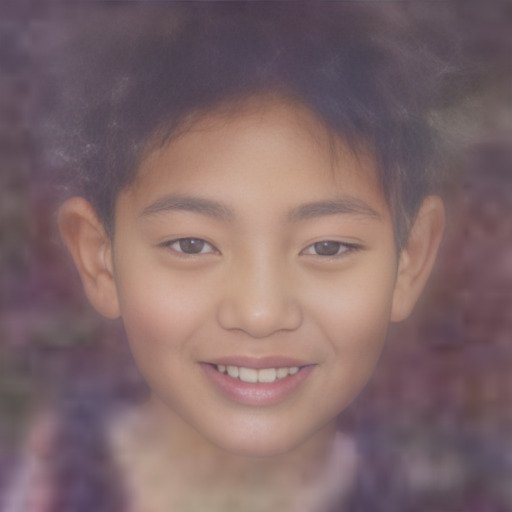} &
\includegraphics[width=0.1\linewidth]{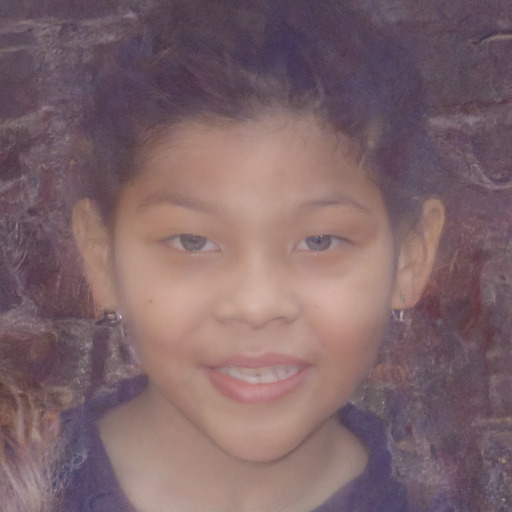} &
\includegraphics[width=0.1\linewidth]{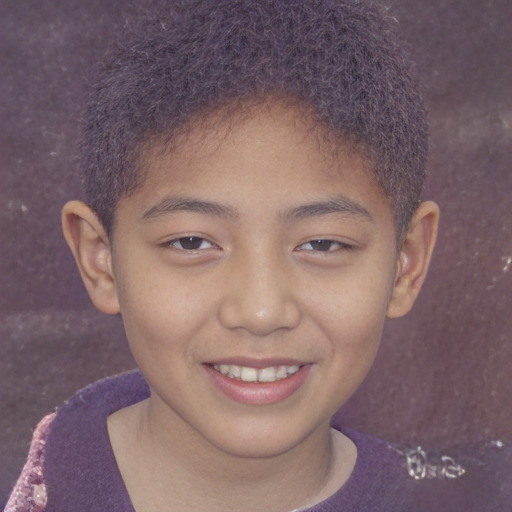} &
\includegraphics[width=0.1\linewidth]{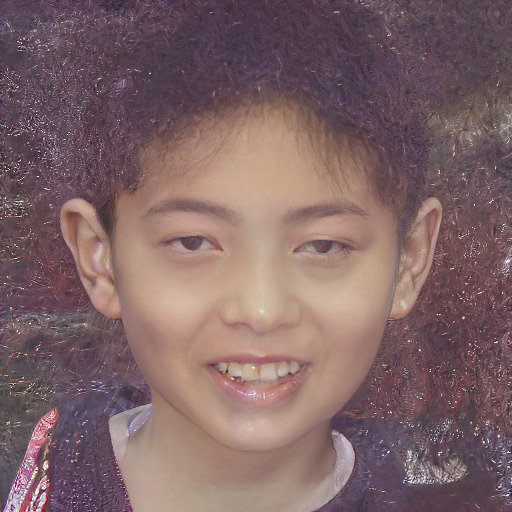} &
\includegraphics[width=0.1\linewidth]{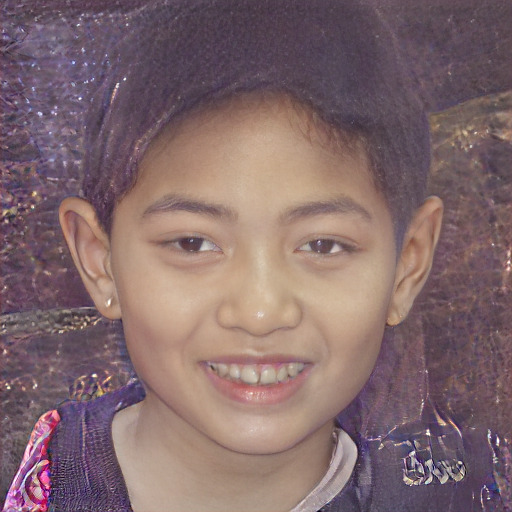} &
\includegraphics[width=0.1\linewidth]{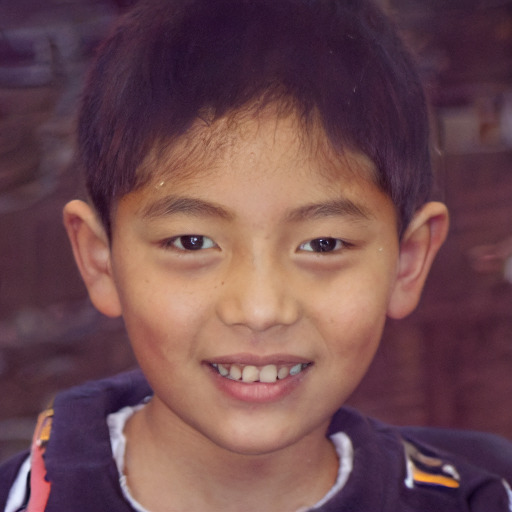} 
\\
Input & GPEN & GFPGAN & DifFace & CodeFormer & RestoreFormer & VQFR & Ours
\end{tabular}\\ %\hspace{-2.3mm}

\caption{Qualitative comparisons on real-world datasets. The first two rows represent images from WIDER face dataset, the  third row represents images from WebPhoto, respectively.}
\label{fig:celeba_sr}
\end{figure*}

\begin{figure*}[t]
\setlength{\tabcolsep}{1pt}
\scriptsize
\centering
%\hspace{-0.4cm}
\begin{tabular}{cccc|cccc}
\includegraphics[width=0.1\linewidth]{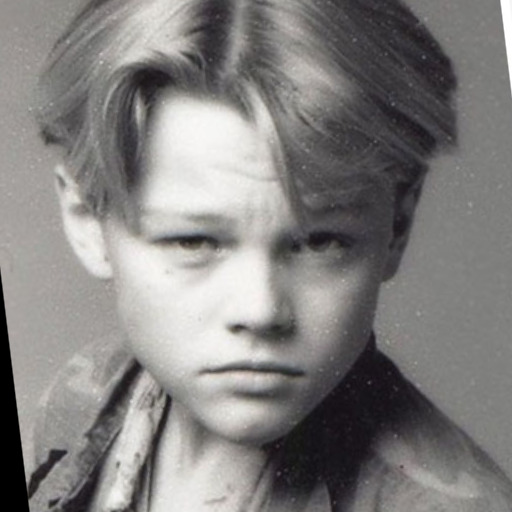} &
\includegraphics[width=0.1\linewidth]{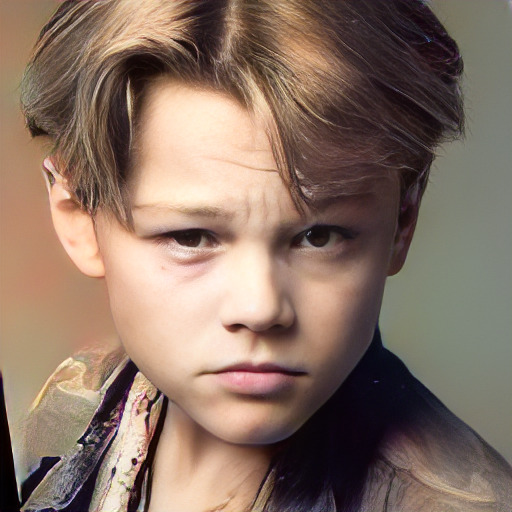} &
\includegraphics[width=0.1\linewidth]{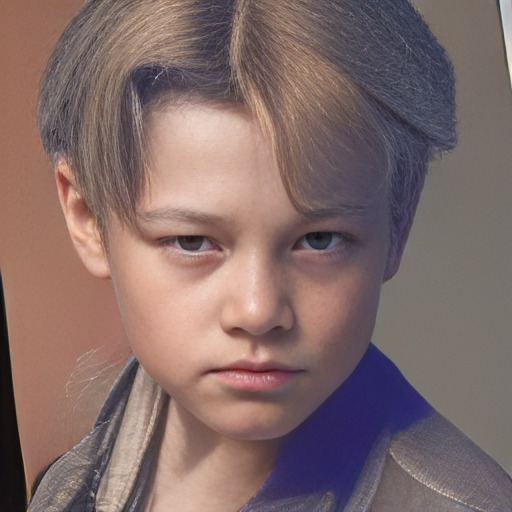} &
\includegraphics[width=0.1\linewidth]{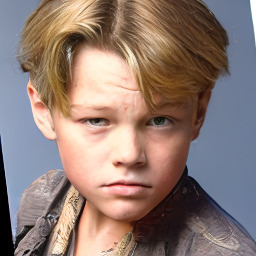} &
\includegraphics[width=0.1\linewidth]{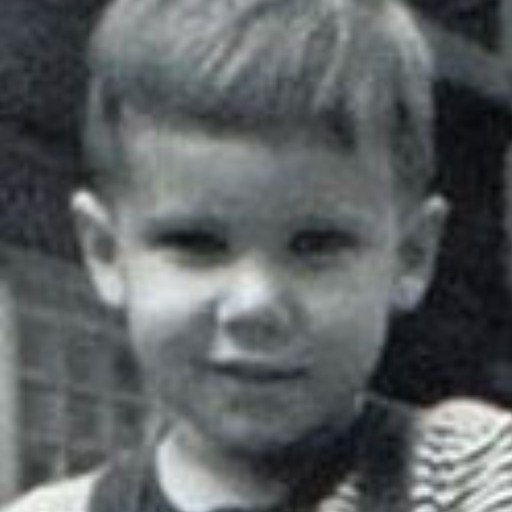} &
\includegraphics[width=0.1\linewidth]{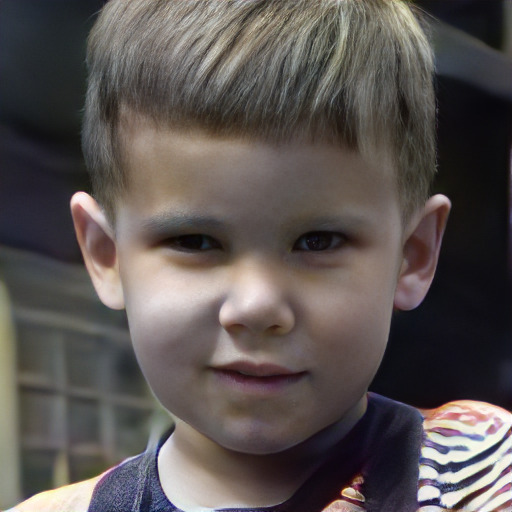} &
\includegraphics[width=0.1\linewidth]{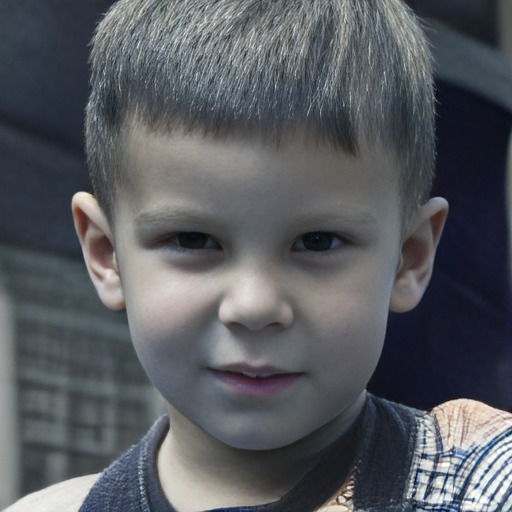} &
\includegraphics[width=0.1\linewidth]{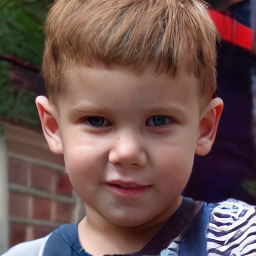} \\
Input & GFPGAN & CodeFormer & Ours & Input & GFPGAN & CodeFormer & Ours
\end{tabular}\\ %\hspace{-2.3mm}

\caption{Qualitative comparisons on real-world images from CelebA-Child for image colorization.}
\label{fig:color}
\end{figure*}
% \vspace{-8mm}

\subsection{Identity Recovery Network (IRN)}
Existing works typically introduce an identity-preserving loss to the same restoration network in addition to standard perceptual losses. But, often, minimizing both losses together hampers the perceptual quality and may introduce unwanted artifacts, especially if the coefficient for identity loss is high. Instead, to ease the learning process, we aim to disentangle these two objectives and use a separate IRN ($f_\phi$) to focus on recovering the identity. We train $f_\phi$, emphasizing the identity-preserving loss, with a small amount of standard $L_1$ loss for better stability. The loss function for training $f_\phi$ is as follows:
\begin{equation}
\label{eq:cen}
    \mathcal{L}_{IRN} = \alpha L_1(f_\phi(\textbf{x}^d), \textbf{x}) + D_{cos}(f_{arc}(f_\phi(\textbf{x}^d)), f_{arc}(\textbf{x}))
\end{equation}
where $D_{cos}$ denotes the cosine distance between two feature vectors and $f_{arc}$ denotes a pre-trained ArcFace model. We denote the second term in Eq. \ref{eq:cen} as  $\mathcal{L}_{arc}$. The IRN aims to generate a stable approximation $\textbf{x}^{id}$ that primarily recovers the identity information rather than focusing on visually pleasing sharp results. Such crucial identity features are also transferred to the latent embedding $\textbf{z}^{id}$. To ensure that during reverse diffusion process, the output does not deviate from the already recovered identity in $\textbf{x}^{id}$, we add a gradient term $\nabla_{{\hat{\textbf{z}}}_{0,t}} \mathcal{L}_{arc}$ to the estimated score at each step, where the $\mathcal{L}_{arc}$ is calculated between the image obtained after projecting $\hat{\textbf{z}}_{0,t}$ to pixel-space using the pretrained decoder and $\textbf{x}^{id}$, $\hat{\textbf{z}}_{0,t}$ is the denoised output at step $t$. It follows the intuition of classifier guidance in \cite{dhariwal2021diffusion}. But, we have observed that simply adding the gradient for the whole latent again deteriorates the perceptual quality. We hypothesize that the latent embedding should have specific features that correlate to identity-specific details in the pixel space. We should ideally regularize the score function of only those features while keeping the rest untouched.

To verify this, we introduce a learnable latent mask $M_l \in \mathbb{R}^{d \times \frac{H}{f} \times \frac{W}{f}}$, denoting the identity-specific latent features. We use a stack of a few convolutional layers followed by a sigmoid layer as $f_M$, that takes the $\hat{\textbf{z}}^c_{t}$ as input and produces $M_l$. To train  $f_M$ , we first update the score to obtain the corresponding $\hat{\textbf{z}}_{0,t}$ and project it back to pixel space. We use LPIPS, cosine-distance-based ID loss on the output image, and a sparsity constraint on $M_l$ to ensure $M_l$ selects only the subset of identity-specific features without harming the LPIPS score. As discussed in our ablation study, the observed improvement after introducing the latent mask validates our design choice. For all the cases, straight-through gradient estimator \cite{esser2021taming} is utilized to copy the gradients from the decoder to the continuous latent embedding to handle the non-differentiable quantization operation.

\section{Experimental Results}
\begin{table}
\begin{center}
    \caption{Quantitative evaluation on 3000 images of size $512 \times 512$ from the CelebA-Test (BFR). Bold and underline indicate the best and the second
best performance.}
    \label{tab:celeba_bfr_512}
    \resizebox{0.4\textwidth}{!}{% use resizebox with textwidth
    \begin{tabular}{|c|c|c||c|c||c|c|} % <-- Alignments: 1st column left, 2nd middle and 3rd right, with vertical lines in between
    \hline
      Methods & FID $\downarrow$ & LPIPS $\downarrow$ & IDS $\uparrow$ & LDM $\downarrow$ &  PSNR $\uparrow$ & SSIM $\uparrow$ \\
      \hline
      GPEN & 101.12 & 0.3362 & 0.4022 & 10.76  & 22.43 & 0.6009\\
      GFPGAN & 99.03 & 0.2812 & 0.4633 & 10.49  & \underline{22.50} & \underline{0.6060}\\
      PSFRGAN & 64.81 & 0.2513 & 0.3983 & 6.51  & 21.75 & 0.5450\\
      CodeFormer & 54.41 & 0.2288 & \underline{0.5009} & 5.85  & 22.35 & 0.5736\\
      DifFace & \underline{52.18} & \underline{0.2061} & 0.4833 & \underline{5.40}  & \textbf{22.74} & \textbf{0.6116}\\
      RestoreFormer & 59.51 & 0.2899 & 0.4041 & 7.65  & 21.73 & 0.5256\\
    VQFR & 60.75 & 0.3147 & 0.37 & 8.32  & 21.02 & 0.4931\\ 
      \hline
      Ours & \textbf{46.15} & \textbf{0.1868} & \textbf{0.5667} & \textbf{5.00}  & 21.66 & 0.5652\\
      \hline
    \end{tabular}%
    }
      \end{center}
  \end{table}
\subsection{Training Dataset}
The FFHQ dataset \cite{karras2019style} contains 70,000 high-quality face images at $1024 \times 1024$ resolution. We resized the images to $512 \times 512$ for training. We synthesized degraded images on the FFHQ dataset using the degradation model proposed in \cite{yang2021gan,wang2021towards,wang2022restoreformer,chen2021progressive}: $x^{d} = ((x \otimes k) \downarrow_s + n_\sigma)_q$. Here, $x$, $x^{d}$, $k$, $n_\sigma$, $s$, and $q$ are the clean face image, the corresponding degraded image, the blur kernel, the Gaussian noise with standard deviation $\sigma$, downscaling factor, and the JPEG-compression quality factor, respectively. We randomly and uniformly sampled $\sigma$, $s$, and $q$ from [0,20], [1,32], and [30,90], respectively. We evaluated our approach on 3,000 synthetic images from CelebA-Test dataset \cite{karras2017progressive}. The degraded images in this dataset were synthesized using the same degradation range as our training. Additionally, we tested our method on real-world datasets: WebPhoto-Test \cite{wang2021towards} (407 images), WIDER Face \cite{yang2016wider} (970 images), Celeb-Child \cite{wang2021towards} (180 images) and TURB. For the TURB dataset, we randomly sampled 139 images from the BRIAR \cite{cornett2023expanding} and LRFID \cite{miller2019data} datasets, which provided a more challenging scenario as our models were not trained on severe turbulence-affected images.
\subsection{Evaluation Metrics}
To assess the quality of the restored images quantitatively, we primarily rely on the Frechet Inception Distances (FID) \cite{heusel2017gans} and Learned Perceptual Image Patch Similarity (LPIPS) \cite{zhang2018unreasonable} metrics. We also compute PSNR and SSIM for completeness, although they often fail to capture visual quality. To evaluate the face recognition performance of the restored images, we calculate the cosine similarity between the features of the restored image and the paired GT image (IDS). Higher cosine similarity values indicate better recovery of the identity information. We use the same evaluation metrics and pre-trained models for LPIPS, FID, and ArcFace as used in the prior art \cite{wang2022restoreformer,wang2021towards}. We also adopt landmark distance (LMD) as the fidelity metric, following \cite{gu2022vqfr}.
\begin{table}
    \centering
    \caption{Quantitative evaluation on 3000 images from the CelebA-Test for extreme upsampling ($16 \rightarrow 512$). Bold and underline indicate the best and the second
best performance.}
    \label{tab:sr}
    \resizebox{0.4\textwidth}{!}{% use resizebox with textwidth
    \begin{tabular}{|c|c|c||c|c||c|c|} % <-- Alignments: 1st column left, 2nd middle and 3rd right, with vertical lines in between
    \hline
      Methods & LPIPS $\downarrow$ & FID $\downarrow$ & IDS $\uparrow$ & LDM $\downarrow$ & PSNR $\uparrow$ & SSIM $\uparrow$ \\
      \hline
      GPEN & 0.4350 & 148.39 & 0.1843 & 21.80 & {19.91} & \underline{0.5346}\\
      GFPGAN & 0.4028 & 160.29 & 0.2243 & 23.10 & \underline{19.95} & \textbf{0.5366}\\
      CodeFormer & {0.3565} & {73.45} & {0.2546} & {11.24} & 19.14 & 0.4639\\
      DifFace & \underline{0.3001} & \underline{53.93} & \underline{0.2892} & \underline{8.85} & \textbf{20.12} & {0.5314} \\
      RestoreFormer & 0.4193 & 103.13 & 0.1438 & 14.10 & 19.26 & 0.4581 \\
      VQFR & 0.4090 & 109.97 & 0.1583 & 15.30 & 18.88 & 0.4220 \\
      Ours & \textbf{0.2574} & \textbf{39.71} & \textbf{0.3512} & \textbf{7.67} & 18.77 & 0.4751 \\
      \hline
      
    \end{tabular}%
    }
  \end{table}
\begin{table}
  \begin{center}
    \caption{Quantitative comparisons of FID ($\downarrow$) on real-world datasets in
terms of FID. Bold and underline indicate the best and the second
best performance, respectively.}
    \label{tab:real}
    \resizebox{0.35\textwidth}{!}{% use resizebox with textwidth
    \begin{tabular}{|c|c|c|c|c|} % <-- Alignments: 1st column left, 2nd middle and 3rd right, with vertical lines in between
    \hline
      Methods & WIDER Face & WebPhoto & CelebA-Child  \\
      \hline
      % Pulse & 80.46 & 87.16 & \textbf{104.80} & - & - \\
      PSFRGAN & 49.85 & 88.45 & {107.40}  \\
      GPEN & 46.99 & 81.77 & 109.55  \\
      GFPGAN & 39.76 & 87.35 & 111.78  \\
      CodeFormer & 39.21 & {78.87}  & 116.18 \\
      VQFR & {44.54} & \underline{75.46} & \underline{105.68}  \\
      DifFace & \underline{37.49} & 85.52 & 110.81 \\
      Ours & \textbf{34.25} & \textbf{75.05} & \textbf{104.40}  \\
      \hline
    \end{tabular}%
    }
      \end{center}
  \end{table}
\subsection{Comparisons with State-of-the-Art Methods}
We compare with the following state-of-the-art (SOTA) methods: PSFRGAN \cite{chen2021progressive}, GPEN \cite{yang2021gan}, GFPGAN \cite{wang2021towards}, CodeFormer \cite{zhou2022towards}, RestoreFormer \cite{wang2022restoreformer}, DifFace \cite{yue2022difface} and VQFR \cite{gu2022vqfr}. We use the official results and checkpoints provided by the authors.
\newline \textbf{Synthetic BFR:} We first evaluated restoration accuracy on the synthetic CelebA-Test dataset for BFR task.  Our approach achieves a much better balance between fidelity (IDS, LDM) and perceptual metrics (LPIPS, FID) as mentioned in Table \ref{tab:celeba_bfr_512}. 
%%%%%%%%%%%%%%%%%%%%%%%%%%%%%%%%%%%%%%%%%%%%%%%%%%%%
\begin{figure*}[t]
\hspace{35pt}
\parbox{.2\linewidth}{
\setlength{\tabcolsep}{1pt}
\scriptsize
\centering
\begin{tabular}{ccc}
\includegraphics[trim={5cm 8cm 3cm 0cm},clip,width=0.36\linewidth]{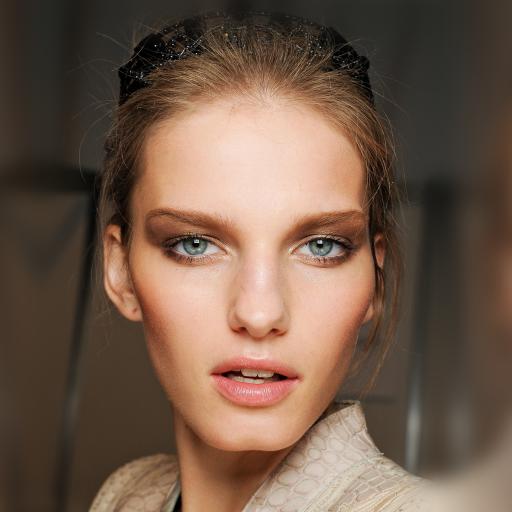} &
\includegraphics[trim={5cm 8cm 3cm 0cm},clip,width=0.36\linewidth]{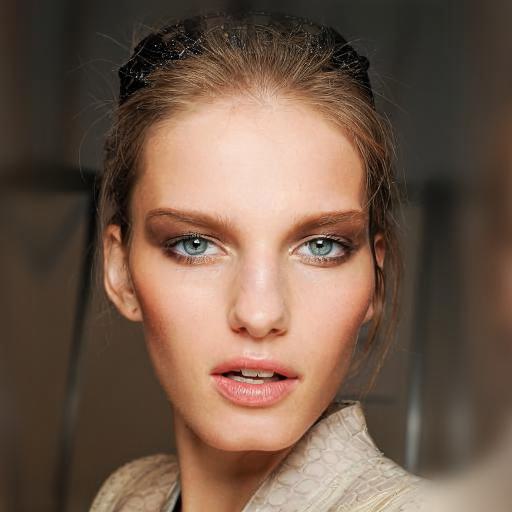} &
\includegraphics[trim={5cm 8cm 3cm 0cm},clip,width=0.36\linewidth]{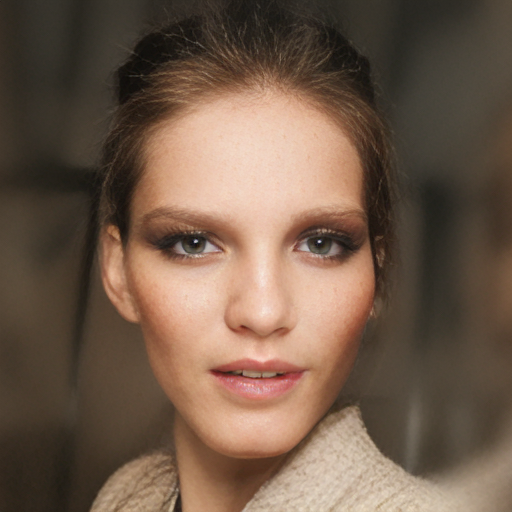} \\
GT & VQVAE (f=4) & VQVAE (f=32) 
\end{tabular} %\hspace{-2.3mm}
}\hspace{20pt}
\parbox{.6\linewidth}{
\setlength{\tabcolsep}{1pt}
\scriptsize
\centering
\begin{tabular}{cccccc}
\includegraphics[width=0.15\linewidth]{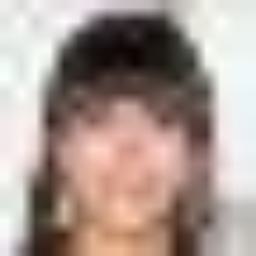} &
\includegraphics[width=0.15\linewidth]{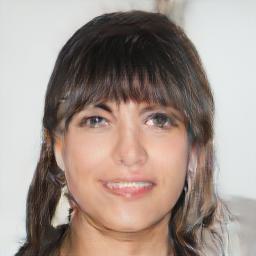} &
\includegraphics[width=0.15\linewidth]{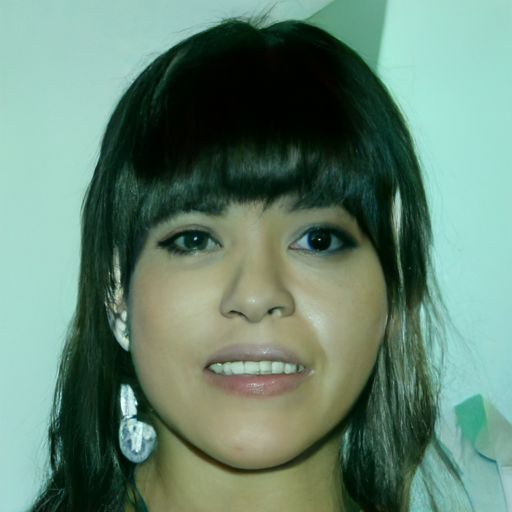} &
\includegraphics[width=0.15\linewidth]{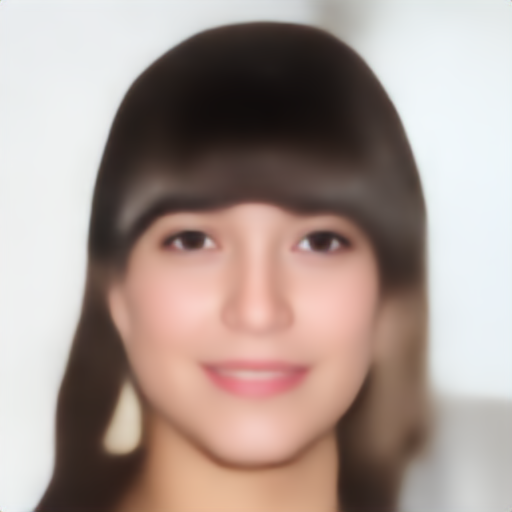} &
\includegraphics[width=0.15\linewidth]{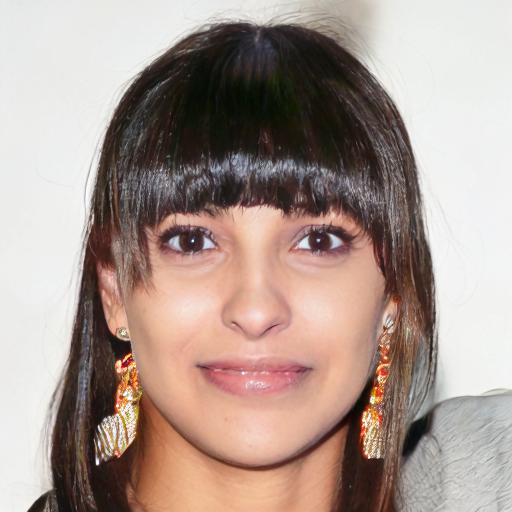} &
\includegraphics[width=0.15\linewidth]{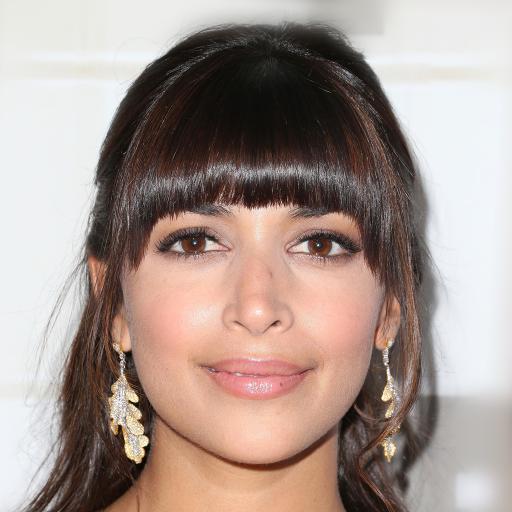} \\ 
IP & VQVAE-Rest & GD (t=1000) & IRN & Ours & GT 
\end{tabular}%\hspace{-2.3mm}
% \caption{Restoration performance of VQVAE, CEN and our final model using latent space refinement. }
}
\caption{(Left): Reconstruction performance of VQVAE. (Right): Restoration performance of VQVAE, GD, IRN and our final model using latent space refinement. }
\label{fig:abl_vq}
\end{figure*}
\newline \textbf{Extreme Upscaling:} We tested the algorithms under extreme BFR conditions, where we applied a fixed downscaling factor of $\times 32$ to $512 \times 512$ images, resulting in degraded images of size $16 \times 16$. To make the task more challenging, we added noise and blur to the images. The quantitative results can be found in Table \ref{tab:sr}. Despite the limited information in the input, our approach outperforms other algorithms in terms of perceptual quality and recognition accuracy.
\newline \textbf{Real-World Cases:} Our approach outperforms existing GAN-based methods like GPEN and GFPGAN in terms of producing more realistic and faithful reconstructions on real-world degraded datasets. CodeForm, VQFR's outputs show visible artifacts or repetitive skin/hair texture due to highly compressed latent space. DifFace outputs are less sharp and may alter facial details as the underlying model is unconditional. In contrast, our approach achieves superior results with fewer artifacts, even for low to medium degradation levels. Real-world colorization examples are shown in Fig. \ref{fig:color}, for which we finetune our network for colorization. We also report the accuracy for the downstream face-recognition task on the TURB dataset in Table \ref{tab:recog}. Our approach comfortably outperforms prior arts for recognition, as well.

\section{Ablation Analysis}
%%%%%%%%%%%%%%%%%%%%%%%%%%%%%%%%%%%%%%%%%%%%%%%%%%%%%%
\begin{table}
  \centering
  \caption{Quantitative comparison of different ablations of our network on CelebA-Test set. ID represents: ArcFace based identity loss. The first two-row represents reconstruction of GT images using VQ-VAE and GAN.}
    \label{tab:ablation}
    \resizebox{0.4\textwidth}{!}{% use resizebox with textwidth
    \begin{tabular}{|c|c|c|c|c|} % <-- Alignments: 1st column left, 2nd middle and 3rd right, with vertical lines in between
    \hline
      Methods & LPIPS $\downarrow$ & FID $\downarrow$ & IDS $\uparrow$ & PSNR $\uparrow$ \\
      \hline
      VQVAE-Rec (f=32) & 0.0559 & 22.76 & 00.9012 & 29.46\\
      VQVAE-Rec (f=4) & \textbf{0.0068} & \textbf{9.02} & \textbf{0.9903} & \textbf{35.72}\\
      \hline
      VQVAE-Rest (f=4) & 0.2127 & 81.69 & 0.5258 & 19.09\\
      IRN & 0.3344 & 127.53 & 0.5155 & \textbf{24.26} \\
      IRN$^{ID}$& 0.3109  & 118.50 & \textbf{0.5849} & 24.16\\
      Diff. Prior & 0.1901 & 47.91 & 0.5107 & 21.06 \\
      IRN$^{ID}$ + Diff. Prior & 0.2004 & 53.22 & 0.5712 & 22.34\\
      IRN$^{ID}$ + Diff. Prior + $M_l$ & \textbf{0.1868} & \textbf{46.15} & 0.5667 & 21.66\\
      \hline
    \end{tabular}%
    }
  \end{table}

In Table \ref{tab:ablation}, we analyze the effect of individual components of our approach on the perceptual quality and identity-preserving aspect. We use CelebA-Test with $512 \times 512$ images for our ablation. We empirically found that the IRN is model agnostic, and any SOTA restoration network results in a comparable accuracy boost. We provided detailed comparisons of possible options for IRN in the supplementary. We finally select SwinIR \cite{liang2021swinir} that keeps a good balance between performance and accuracy. We design our VQ backbone with the following settings: f=4, codebook = 8192, latent = $3 \times 128 \times 128$, feat = 128 ($\times 1, \times 2, \times 3$). For the diffusion model, we have used 40 time steps with heun sampler \cite{karras2022elucidating} and a UNet-based denoiser based on GD \footnote{GD: https://github.com/openai/guided-diffusion}. More implementation and network details, additional results are provided in the supplementary material.
\newline \textbf{Compression Factor}: First, we compared VQVAE models with different latent sizes: $f=32$ and $f=4$. The $f=32$ models, commonly used in previous works \cite{gu2022vqfr}, resulted in poor reconstruction quality due to significant spatial compression to $16 \times 16$. In contrast, the $f=4$ models achieved superior reconstruction quality, as shown in Fig. \ref{fig:abl_vq} (Left). We have empirically verified that it is satisfactorily accurate in reconstructing degraded images as well (PSNR=37.9). Higher spatial compression led to the loss of fine facial details and overly smooth outputs. For example, in the first row, the eye color was changed, and in the second row, both the eye and hair patterns were altered in the $f=32$ models. 
\newline \textbf{Latent Refinement}: We attempted to fine-tune the VQVAE model (with $f=4$) directly for restoration, but it often failed to remove degradation effectively, as reported in \cite{gu2022vqfr}. To address this issue, we developed a modified version called VQVAE-Rest. This variant takes both the degraded image and the coarse estimate from the IRN as inputs, enabling it to focus on recovering residual details. However, even with this modification, the output of VQVAE-Rest still exhibited lower quality and noticeable artifacts (Figure \ref{fig:abl_vq} (Right), Column 2). These results show the need for appropriate latent-space refinement to achieve satisfactory restoration outcomes. We also compare with Guided-Diffusion (GD) for pixel-space-based diffusion (DDPM), which is run for 1000 timesteps in Figure \ref{fig:abl_vq} (Right). As can be observed, it is not only computationally intensive due to large no. of time steps, but also fails to reconstruct the image properly starting from pure noise. The outputs of the IRN shown in Figure \ref{fig:abl_vq} (Right), Column 4, may lack sharp details. However, it recovers crucial facial locations like eye, nose, etc. 
\newline \textbf{ID Loss}: Table \ref{tab:ablation} demonstrates the significance of the ArcFace \cite{deng2019arcface} based identity-preserving loss for the IRN (IRN$^{ID}$ vs. IRN), which prioritizes the recovery of identity-specific features, which are valuable for regularizing the diffusion process. 
\newline \textbf{Adaptive Mask}: While the diffusion model alone can achieve good perceptual quality, it may inadvertently alter certain identity-specific facial features. Adding IRN$^{ID}$-based gradient improves the IDS, but is detrimental for perceptual metrics. The final output could be slightly different from
the output of IRN as we regularize only a subset of the latent in the diffusion stage. Our final model, combining the strengths of the Diff. Prior, IRN$^{ID}$ and the learnable latent mask $M_l$ achieves superior accuracy both in terms of perceptual metrics and fidelity.
\newline Inference time(s) for GFPGAN, CodeFormer, DifFace,
Guided-Diffusion and ours are: 0.5, 0.15, 6,30, 0.61. Our design is still much faster than existing diffusion-based BFR methods, and can potentially be further accelerated by reducing the number of steps using knowledge-distillation techniques \cite{song2023consistency}. We also have comparable or less parameters (16M) compared to VQFR (76.3M), DifFace (17.5), RestoreFormer (12M).
%%%%%%%%%%%%%%%%%%%%%%%%%%%%%%%%%%%%%%%%%%%%%%%%%%%%%%
\begin{table}
  \begin{center}
  \caption{Face recognition accuracy using pre-trained Arcface on real-world BRIAR and\textit{LRFID} dataset. Our method performs best for such downstream task as well.}
    \label{tab:recog}
    \resizebox{0.3\textwidth}{!}{% use resizebox with textwidth
    \begin{tabular}{|c|c|c|c|} % <-- Alignments: 1st column left, 2nd middle and 3rd right, with vertical lines in between
    \hline
      Methods & Top-1 ($\uparrow$) & Top-3 ($\uparrow$) & Top-5 ($\uparrow$) \\
      \hline
      GPEN & 32/\textit{50.6 }& 50/\textit{72} & 62/\textit{81} \\
      GFPGAN & 26/\textit{57} & 58/\textit{79} & 60/\textit{85}\\
      CodeFormer & 28/\textit{61} & 52/\textit{77}  & 58/\textit{82}\\
      RestoreFormer & 20/\textit{\underline{62}} & 50/\textit{\underline{81}}  & 62/\textit{\underline{88}}\\
      VQFR & \underline{34}/\textit{60} & \underline{60}/\textit{79} & \underline{64}/\textit{87} \\
      Ours & \textbf{34/\textit{68}} & \textbf{68/\textit{81}} & \textbf{72/\textit{90}} \\
      \hline
      
    \end{tabular}%
    }
      \end{center}
  \end{table}

\vspace{-3mm}
\section{Conclusions}
% \begin{figure}[h]
% %\newlength\fsdurthree
% %\setlength{\fsdurthree}{-1.5mm}
% \setlength{\tabcolsep}{1pt}
% \scriptsize
% \centering
% %\hspace{-0.4cm}
% \begin{tabular}{ccc|ccc}
% % \includegraphics[width=0.095\linewidth]{limit/set1/02744_ip.jpg} &
% % \includegraphics[width=0.095\linewidth]{limit/set1/02744_00_vqfr.png} &
% % \includegraphics[width=0.095\linewidth]{limit/set1/02744_ours.jpg} &
% \includegraphics[width=0.13\linewidth]{limit/set2/02839_ip.jpg} &
% \includegraphics[width=0.13\linewidth]{limit/set2/02839_00_vqfr.png} &
% \includegraphics[width=0.13\linewidth]{limit/set2/02839_ours.jpg} &
% \includegraphics[width=0.13\linewidth]{limit/set3/02864_ip.jpg} &
% \includegraphics[width=0.13\linewidth]{limit/set3/02864_00_vqfr.png} &
% \includegraphics[width=0.13\linewidth]{limit/set3/02864_ours.jpg}
% \\
% Input & VQFR & Ours & Input & VQFR & Ours
% \end{tabular}\\ %\hspace{-2.3mm}
% \vspace{-3mm}
% \caption{Limitations of our approach on challenging cases.}
% \vspace{-5mm}
% \label{fig:limit}
% \end{figure}
We propose a iterative latent-space refinement technique using diffusion prior for restoring severely degraded face images, achieving a better balance between restoration quality and identity recovery compared to existing methods. However, the performance of our approach is limited by the IRN's ability to preserve identity information. To improve efficiency, we plan to explore the development of a recognition-model-free reverse algorithm in future research.
\section{Acknowledgement}
This research is based upon work supported in part by the Office of the Director of National Intelligence (ODNI), Intelligence Advanced Research Projects Activity (IARPA), via [2022-21102100005]. The views and conclusions contained herein are those of the authors and should not be interpreted as necessarily representing the official policies, either expressed or implied, of ODNI, IARPA, or the U. S. Government. The US. Government is authorized to reproduce and distribute reprints for governmental purposes notwithstanding any copyright annotation therein.

\bibliographystyle{named}
\bibliography{ijcai24}

\end{document}